\documentclass[10pt,twocolumn,letterpaper]{article}

\usepackage[pagenumbers]{iccv} % To force page numbers, e.g. for an arXiv version

\usepackage{soul}
\usepackage{textcomp,gensymb}
\usepackage{multirow}
\usepackage{booktabs}    % For professional-quality tables
\usepackage{adjustbox}   % To adjust table size
\usepackage{caption}
\usepackage{float}
\usepackage{placeins}
\usepackage{amssymb}
\usepackage{pifont}
\usepackage{placeins} % Add in the preamble if not already included
\usepackage{arydshln}
\usepackage[accsupp]{axessibility}
\definecolor{DarkGreen}{RGB}{0,100,0}
\definecolor{DarkRed}{RGB}{139,0,0}

\newcommand{\inc}[1]{{\color{DarkGreen} $\uparrow$ #1}}
\newcommand{\dec}[1]{{\color{DarkRed} $\downarrow$ #1}}
\newcommand{\decreg}[1]{{\color{DarkGreen} $\downarrow$ #1}}

\definecolor{iccvblue}{rgb}{0.21,0.49,0.74}
\usepackage[pagebackref,breaklinks,colorlinks,allcolors=iccvblue]{hyperref}

\def\paperID{9777} % *** Enter the Paper ID here
\def\confName{ICCV}
\def\confYear{2025}

\title{Towards a Unified Copernicus Foundation Model for Earth Vision}

\author{
    Yi Wang\textsuperscript{1} \qquad
    Zhitong Xiong\textsuperscript{1} \qquad
    Chenying Liu\textsuperscript{1,2} \qquad
    Adam J. Stewart\textsuperscript{1,2} \qquad
    Thomas Dujardin\textsuperscript{1} \qquad \\
    Nikolaos Ioannis Bountos\textsuperscript{3,4} \qquad
    Angelos Zavras\textsuperscript{3,4} \qquad
    Franziska Gerken\textsuperscript{5} \qquad \\
    Ioannis Papoutsis\textsuperscript{3} \qquad
    Laura Leal-Taixé\textsuperscript{5} \qquad
    Xiao Xiang Zhu~\textsuperscript{1,2} \\
    \textsuperscript{1} Technical University of Munich \qquad
    \textsuperscript{2} Munich Center for Machine Learning \qquad \\
    \textsuperscript{3} National Technical University of Athens \& National Observatory of Athens \qquad \\
    \textsuperscript{4} Harokopio University of Athens \qquad \textsuperscript{5} NVIDIA \\
}

\begin{document}
\maketitle

% \begin{itemize}
%     \item \red{If sidebar comment not available, use \textbackslash red\{\} or create new commands in \texttt{preamble.tex} to add comments.}
%     \item \red{Track change will also be turned on.}
    
% \end{itemize}

\begin{abstract}
Advances in Earth observation (EO) foundation models have unlocked the potential of big satellite data to learn generic representations from space, benefiting a wide range of downstream applications crucial to our planet.
However, most existing efforts remain limited to fixed spectral sensors, focus solely on the Earth's surface, and overlook valuable metadata beyond imagery.
In this work, we take a step towards next-generation EO foundation models with three key components: 1) Copernicus-Pretrain, a massive-scale pretraining dataset that integrates 18.7M aligned images from all major Copernicus Sentinel missions, spanning from the Earth's surface to its atmosphere; 2) Copernicus-FM, a unified foundation model capable of processing any spectral or non-spectral sensor modality using extended dynamic hypernetworks and flexible metadata encoding; and 3) Copernicus-Bench, a systematic evaluation benchmark with 15 hierarchical downstream tasks ranging from preprocessing to specialized applications for each Sentinel mission. Our dataset, model, and benchmark greatly improve the scalability, versatility, and multimodal adaptability of EO foundation models, while also creating new opportunities to connect EO, weather, and climate research. Codes at https://github.com/zhu-xlab/Copernicus-FM.
\end{abstract}    
\vspace{-1.5em}
\section{Introduction}
\label{sec:intro}

% An extension of abstract, more texts on background, our motivation, and contributions.

% Motivation-1: existing EO FM research (data) is limited to medium-to-high resolution, commonly seen RGB, SAR, and optical sensors observing the Earth's surface. ---- We extend the coverage to all major Sentinel missions, including low resolution sensors with wider and more frequent surface coverage, as well as the Earth's atmosphere.

% Motivation-2: existing EO FM research (model) is limited to a few fixed sensors with spectral response. ---- We extend the model's capacity to process any spectral or non-spectral sensors.

% Motivation-3: existing EO FM research (benchmark) is also limited to a few fixed spectral sensors observing the Earth's surface. ---- We provide a comprehensive benchmark with hierarchical tasks covering all major Sentinel missions. 

% Motivation-4: Our dataset naturally aligns between EO and weather/climate, providing opportunities to connect both by using EO features to help weather forecast / climate prediction or vice versa.

Earth observation (EO) satellites provide a critical means of monitoring planetary dynamics, capturing land cover changes, atmospheric composition, and other important environmental phenomena~\cite{zhu2017deep,reichstein2019deep,xiong2024earthnets}. Recent advances in self-supervised learning have revolutionized the usage of big EO data, enabling the development of general-purpose foundation models to learn generalized representations from unlabeled imagery at scale~\cite{wang2022self,zhu2024foundations}. Despite rapid progress, the evolution of EO foundation models faces critical limitations in three dimensions: sensor diversity, model flexibility, and evaluation breadth.

\begin{figure}
    \centering
    \includegraphics[width=0.9\linewidth]{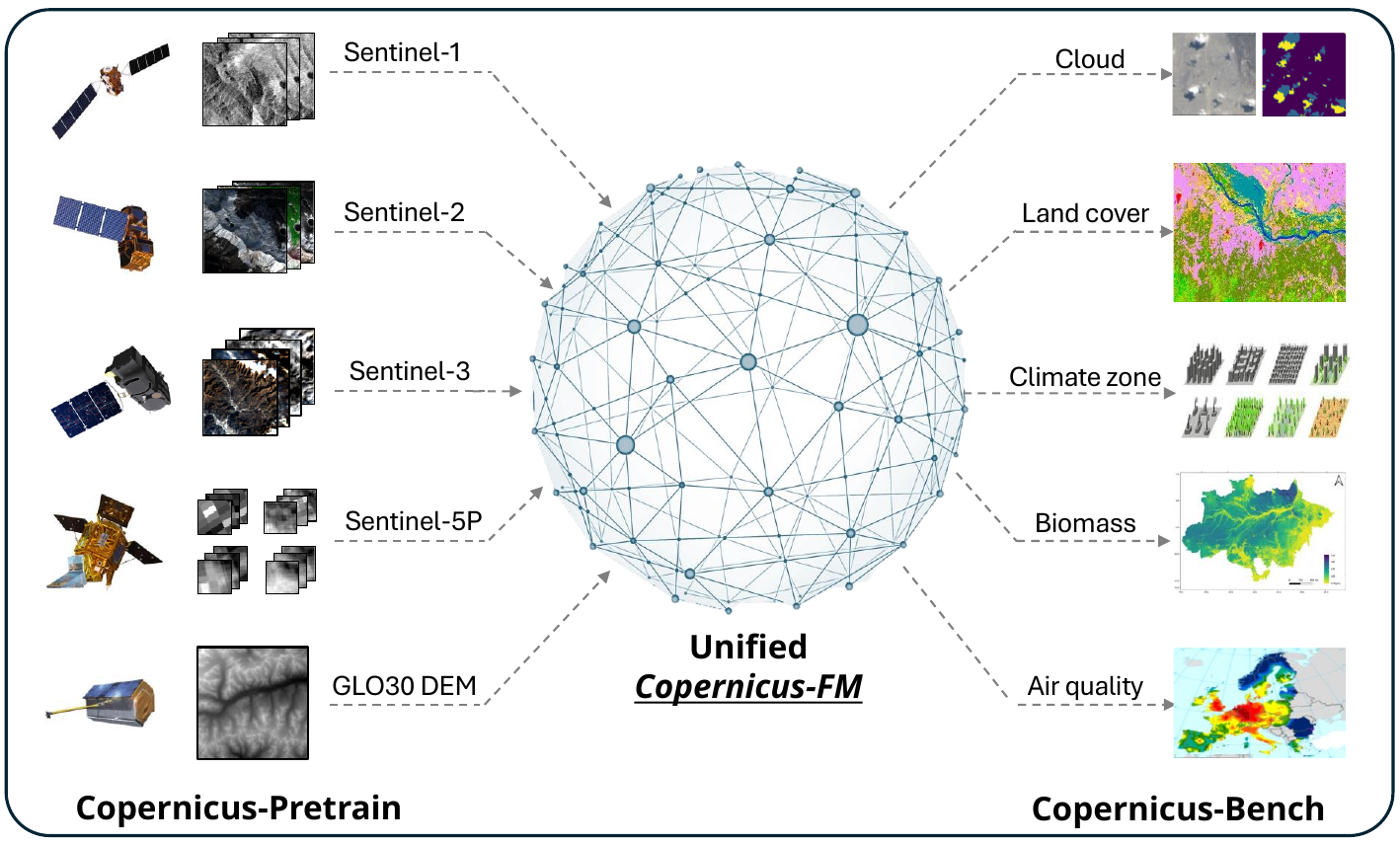}
    \caption{Overview of our efforts towards a unified Copernicus foundation model, from pretraining to benchmarking.}
    \vspace{-1em}
    \label{fig:enter-label}
    
\end{figure}

First, current pretraining datasets predominantly focus on high- to medium-resolution sensors like Sentinel-1/2~\cite{torres2012gmes,drusch2012sentinel} and Landsat~\cite{wulder2022fifty} observing the Earth's surface~\cite{wang2023ssl4eo,stewart2023ssl4eo,bastani2023satlaspretrain}. This excludes lower-resolution but temporally rich missions like Sentinel-3~\cite{donlon2012global} and Sentinel-5P~\cite{veefkind2012tropomi}, which provide near-daily global coverage of land, oceanic, and atmospheric variables critical for climate studies.
Second, most existing EO foundation models adopt rigid architectures tailored to one or more specific sensor modalities, lacking the capacity to dynamically adapt to new spectral bands or non-spectral input. While recent efforts~\cite{xiong2024neural,bountos2023fomo} introduce some spectral flexibility, they still lack mechanisms to handle non-spectral variables that possess a significant amount of EO data. 
Furthermore, current foundation model evaluation benchmarks primarily focus on surface applications using RGB, multispectral, or SAR sensors~\cite{lacoste2023geo,marsocci2024pangaea}, while overlooking coarse-scale sensors and atmospheric tasks. 
%These limitations hinder the development of versatile multimodal foundation models capable of bridging EO with weather and climate research—a pressing need in the era of global environmental change.
%
% While EO monitors the Earth's surface, atmosphere, and ecosystems using remote sensing, the related field of weather and climate research focuses specifically on atmospheric conditions.
%
% EO provides critical data for weather and climate analysis; however these limitations hinder the development of versatile multimodal foundation models capable of bridging EO with weather and climate research--highlighting a promosing opportunity for advancement and an urgent necessity in the era of global environmental change. 
These limitations hinder the development of versatile multimodal foundation models that integrate EO with weather and climate research, highlighting both a critical challenge and a promising opportunity in the era of global environmental change.

To address these challenges, we introduce three synergistic contributions that enhance the scalability, versatility, and multimodal integration of EO foundation models. 
First, we introduce \textbf{Copernicus-Pretrain}, one of the largest and most diverse EO pretraining datasets to date, comprising 18.7 million aligned observations from all major operational Copernicus Sentinel missions (Sentinel-1 to Sentinel-5P). Unlike prior datasets focusing on fine-grained surface observations, Copernicus-Pretrain enables holistic modeling of Earth system interactions by integrating atmospheric variables and coarse-scale observations with wider and more frequent coverage.
Second, we propose \textbf{Copernicus-FM}, a unified foundation model capable of processing any spectral or non-spectral sensor using dynamic hypernetworks. With additional support for metadata integration, it is valuable for a wide range of practical applications. Third, we establish \textbf{Copernicus-Bench}, a comprehensive evaluation benchmark with 15 downstream tasks hierarchically organized across preprocessing (e.g., cloud removal), base applications (e.g., land cover classification), and specialized applications (e.g., air quality estimation). This benchmark enables systematic assessment of foundation model performances across various Sentinel missions on different levels of practical applications.

Our work opens new frontiers in three directions: (1) scaling EO pretraining to unified multimodal datasets, breaking the traditional silos between surface and atmospheric observations; (2) verifying that dynamic architectures can overcome sensor heterogeneity, a longstanding challenge in multi-source remote sensing; and (3) providing the first evidence that joint cross-modal pretraining enhances performance on both surface and atmospheric tasks. Furthermore, our efforts create novel opportunities to integrate EO foundation models with weather and climate prediction systems—for instance, using compressed EO embeddings (with rich semantic information) in addition to geographical coordinates to support climate modeling.
%% Maybe include a last section here which summarizes the structure of the paper: The main structure of this paper is as follows: 
% \cref{sec:intro} gives an introduction,
% \cref{sec:relatedwork} reviews prior work, \cref{sec:ssl4eos} introduces the Copernicus-Pretrain dataset, \cref{sec:dofas} proposes the Copernicus-FM model, \cref{sec:sentinelbench} presents the Copernicus-Bench benchmark, \cref{sec:climate} discusses the potential of bridging EO and climate, and \cref{sec:conclusion} concludes the paper. 
% Since the paper covers so much, it might be helpful to have such an overview before one starts to read the paper in detail.

\section{Related work}
\label{sec:relatedwork}

\paragraph{EO pretraining datasets}  
Early efforts like fMoW~\cite{christie2018functional}, Million-AID~\cite{long2021creating}, and SEN12MS~\cite{schmitt2019sen12ms} pioneered large-scale pretraining with supervised datasets before the era of self-supervised learning. SeCo~\cite{manas2021seasonal} introduced Gaussian sampling around populated regions to enrich the landscape diversity with multiseasonal time series, which was further extended by SSL4EO-S12~\cite{wang2023ssl4eo} and SSL4EO-L~\cite{stewart2023ssl4eo} through overlap and NaN filtering on Sentinel-1/2 and Landsat series. SatlasPretrain~\cite{bastani2023satlaspretrain} and Major TOM~\cite{francis2024major} boost the dataset sizes through more dense global coverage. Recently, increasing efforts have been spent to broaden sensor diversity such as SpectralEarth~\cite{braham2024spectralearth} for hyperspectral pretraining and MMEarth~\cite{nedungadi2024mmearth} that gathers Sentinel-1/2, elevation, and several EO products together. Our Copernicus-Pretrain dataset aligns all primary Sentinel missions (1--5P) with extended global coverage (e.g., polar regions), enabling joint surface--atmosphere modeling at scale.

\vspace{-0.5em}
\paragraph{EO foundation models}

Single-sensor models dominate early foundation model research, which can be categorized based on pretraining strategies into: 1) contrastive methods like GASSL~\cite{ayush2021geography}, SeCo~\cite{manas2021seasonal}, MATTER~\cite{akiva2022self}, CACo~\cite{mall2023change}, SatMIP~\cite{bourcier2024learning}, etc., 2) masked image modeling (MIM) methods such as SatMAE~\cite{cong2022satmae}, Prithvi~\cite{jakubik2023foundation}, SpectralGPT~\cite{hong2024spectralgpt}, Scale-MAE~\cite{reed2023scale}, and many others~\cite{sun2022ringmo,wang2024feature,tang2023cross,noman2024rethinking,nedungadi2024mmearth,li2024masked}, and 3) hybrid methods like GFM~\cite{mendieta2023towards}, SoftCon~\cite{wang2024multi}, SAR-JEPA~\cite{li2024predicting}, etc. Most of these models focus on optical data. In a trend towards multimodal pretraining, mainstream approaches use either separate encoders (DeCUR~\cite{wang2024decoupling}, CROMA~\cite{fuller2024croma}, SkySense~\cite{guo2024skysense}, etc.) or joint encoders with a few fixed modalities~\cite{wang2022self1,irvin2023usat,xiong2024one,han2024bridging,astruc2024omnisat,zhang20242,astruc2024anysat}. While flexible multimodal architectures like FoMo-Net~\cite{bountos2023fomo}, SenPa-MAE~\cite{prexl2024senpa}, and DOFA~\cite{xiong2024neural} support input with any channels, they are restricted to spectral sensors and ignore metadata. Our Copernicus-FM enhances this approach through dynamic hypernetworks that adapt to arbitrary spectral/non-spectral inputs and support metadata integration, effectively unifying surface and atmospheric data streams.

\vspace{-0.5em}
\paragraph{EO benchmarks}

Existing benchmarks for evaluating EO foundation models vary widely in scope and focus but are, in general, limited in sensor and task diversity. SustainBench~\cite{yeh2021sustainbench} targets sustainable development goals with 15 socioeconomic tasks, GEO-Bench~\cite{lacoste2023geo} standardizes 12 classification/segmentation tasks with a majority focusing on optical imagery, PhilEO Bench~\cite{fibaek2024phileo} consists of three Sentinel-2-derived tasks, and FoMo-Bench~\cite{bountos2023fomo} focuses on forest monitoring with 15 multimodal datasets. Recent efforts like PANGAEA~\cite{marsocci2024pangaea} address geographical bias and model generalizability by aggregating diverse datasets but remain surface-centric. Our Copernicus-Bench fills the gap with 15 hierarchical tasks spanning three application levels, covering all major Sentinel missions and encompassing both surface and atmosphere.

\vspace{-0.3em}
\section{Copernicus-Pretrain}
\label{sec:ssl4eos}

% \begin{table*}[ht]
% \centering
% \caption{Data sources to construct the Copernicus-Pretrain dataset.}
% \label{tab:datasource}
% \begin{tabular}{lccccc}
% \hline
% Data source & Modality & Resolution & Revisit time & \# bands & wavelengths \\ \hline
% S1 GRD~\cite{} & SAR & 10m & ~6 days & 2 & 5.5cm \\
% S2 TOA~\cite{} & multispectral & 10m & ~5 days & 13 & 0.44-2.2um \\
% S3 OLCI~\cite{} & multispectral & 300m & ~2 days & 21 & 0.4-1.02um \\
% S5P NO2/CO/O3/SO2~\cite{} & atmospheric variable & 1km & ~1 day & 1(*4) & - \\
% GLO30 DEM~\cite{} & elevation & 30m & - & 1 & - \\ \hline
% \end{tabular}
% \end{table*}

\begin{table*}[htbp]
\centering
\caption{Copernicus-Pretrain dataset statistics.}
\label{tab:datasetstat}
\scalebox{0.9}{\begin{tabular}{lccccccc}
\toprule
\multicolumn{1}{c}{} & Modality & GSD & Image size & \# Grid cells & \# Patches & \# Timestamps & \# Total images \\ \midrule
Sentinel-1 GRD & SAR & 10~m & 264$\times$264$\times$2 & 247,723 & 1,067,267 & $\sim$4 & 4,227,387 \\
Sentinel-2 TOA & MS & 10~m & 264$\times$264$\times$13 & 247,723 & 1,067,267 & $\sim$4 & 4,218,065 \\
Sentinel-3 OLCI & MS & 300~m & 96$\times$96$\times$21 & 281,375 & 281,375 & $\sim$8 & 2,189,561 \\
Sentinel-5P CO & atmos. & 1~km & 28$\times$28 & 306,097 & 306,097 & 1--12 & 2,104,735 \\
Sentinel-5P NO2 & atmos. & 1~km & 28$\times$28 & 291,449 & 291,449 & 1--12 & 1,752,558 \\
Sentinel-5P SO2 & atmos. & 1~km & 28$\times$28 & 262,259 & 262,259 & 1--12 & 1,366,452 \\
Sentinel-5P O3 & atmos. & 1~km & 28$\times$28 & 306,218 & 306,218 & 1--12 & 2,556,631 \\
Copernicus DEM & elevation & 30~m & 960$\times$960 & 297,665 & 297,665 & 1 & 297,665 \\ \midrule
Copernicus-Pretrain &  &  &  & 312,567 & 3,879,597 &  & 18,713,054 \\ \bottomrule
\end{tabular}}
\end{table*}

\begin{figure*}
    \centering
    \includegraphics[width=0.9\linewidth]{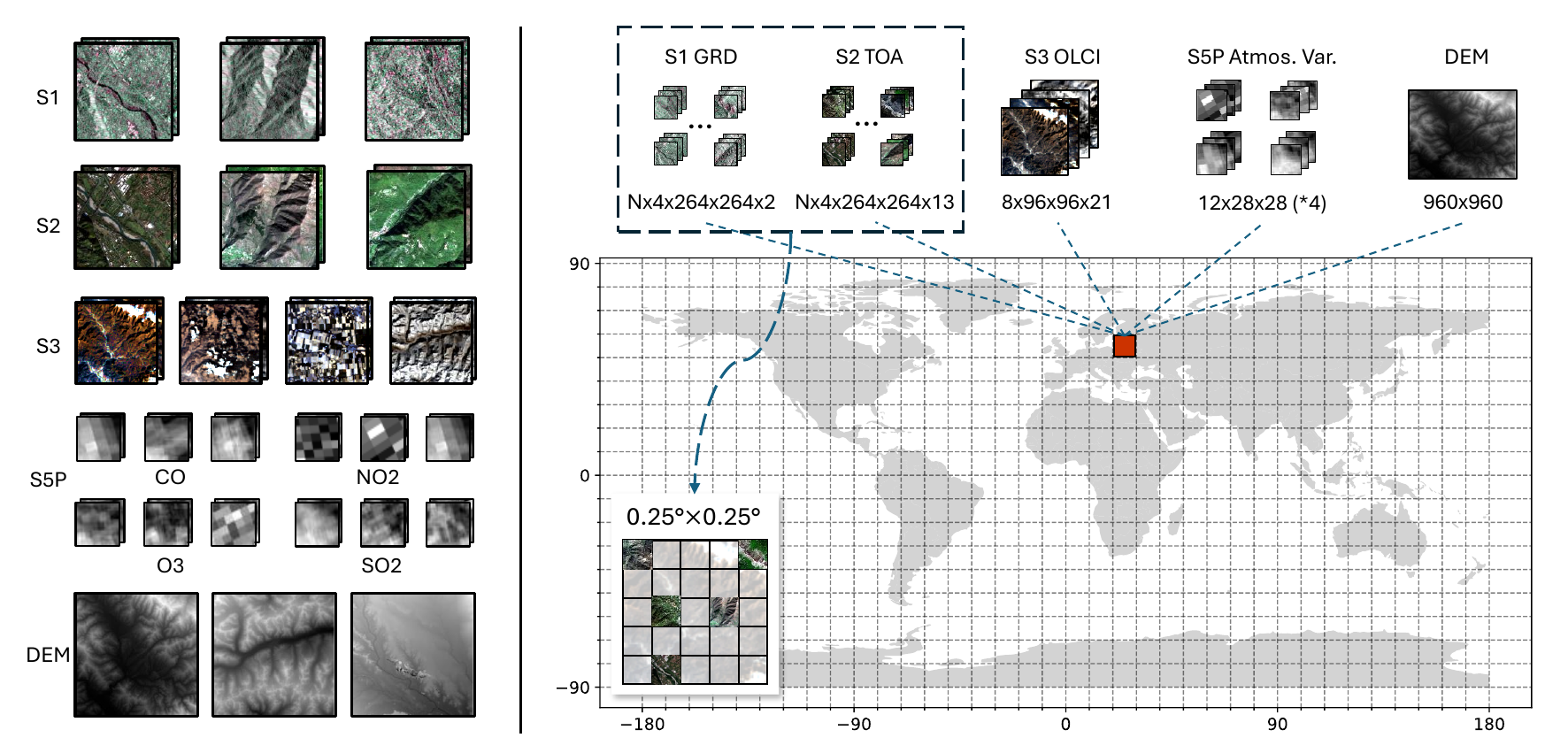}
    \caption{Schematic of the Copernicus-Pretrain dataset. $N$ is the number of local patches. Grid cells are upscaled for ease of visualization.}
    \label{fig:datasetexample}
\end{figure*}

We introduce a unified dataset, Copernicus-Pretrain, containing aligned imagery organized in dense regional grids from all major Sentinel missions in operation (Sentinel-1 SAR, Sentinel-2 multispectral reflectance, Sentinel-3 multispectral radiance, and Sentinel-5P atmospheric variables), as well as an elevation product Copernicus DEM GLO-30~\cite{ESA_Copernicus_2024}. Copernicus-Pretrain significantly extends the scale in data size and modality range of existing EO foundation model research, and provides opportunities for bridging EO and weather/climate studies.

\subsection{Data collection}

The Copernicus-Pretrain dataset is designed for consistency with the ERA5~\cite{hersbach2020era5} reanalysis dataset to provide a convenient alignment between EO imagery and weather/climate data. The workflow of dataset curation is as follows.

First, we divide the globe into $\sim$1M $0.25\degree \times 0.25\degree$ grid cells following the coordinate mapping of ERA5. Each cell thus covers a surface area of about 28~km $\times$ 28~km, forming the basic sample unit of the dataset. With a main interest and data availability in land and its atmosphere, we apply the land mask to filter land grids with a 0.5$\degree$ buffer around the coastline, resulting in about 393K grids.

We then use Google Earth Engine~\cite{gorelick2017google} to download Sentinel images for each grid cell around the anchor year 2021. For Sentinel-3/5P and DEM, we cover the whole cell with patch sizes of about $96\times96$, $28\times28$, and $960\times960$ pixels for each corresponding modality. For Sentinel-3, we filter out cloudy tiles with bright pixel percentages above 20\% and randomly download eight images from the year. For Sentinel-5P, we select four atmospheric variables: NO\textsubscript{2}, CO, SO\textsubscript{2} and O\textsubscript{3}, filter out fully NaN tiles, and download 12 monthly mean images across the year for each variable. It is worth noting that for Sentinel-3 and 5P, it is possible to have fewer images available due to cloud and NaN filtering, thus the final sequence lengths can vary for different grids. For DEM, time series is not applicable, and we simply download one image for the grid.

For Sentinel-1 and 2, the data volume is too large to cover the whole grid cell and the data can be extremely redundant. Therefore, we collect images by sampling local patches within the cells. To optimize the diversity of the data, we do not directly conduct uniform sampling inside a grid, but follow a Gaussian sampling strategy~\cite{manas2021seasonal,wang2023ssl4eo,stewart2023ssl4eo} to sample locations around top-10K populated cities with a standard deviation of 50~km. We utilize this strategy to sample $\sim$1M locations, each downloading 4 seasonal $264\times264$ image patches. We uniformly sample and download images for an additional 40K polar locations. Next, we group these local patches into the predefined $0.25\degree \times 0.25\degree$ grid cells according to their center locations. For those remaining cells that do not contain any local patches, we newly sample and download 1--2 local patches to fill in the gaps.

After the downloading process is finished, a series of quality checks are conducted, removing images that are either corrupted or still dominated by NaN values. In the end, we get $\sim$310K grid cells with at least one sensor available, and $\sim$220K grids with all modalities available. \cref{fig:datasetexample} (right) illustrates an example cell containing all eight modalities.

\subsection{Dataset characteristics}

\begin{figure}[htbp]
    \centering
    \includegraphics[width=0.9\linewidth]{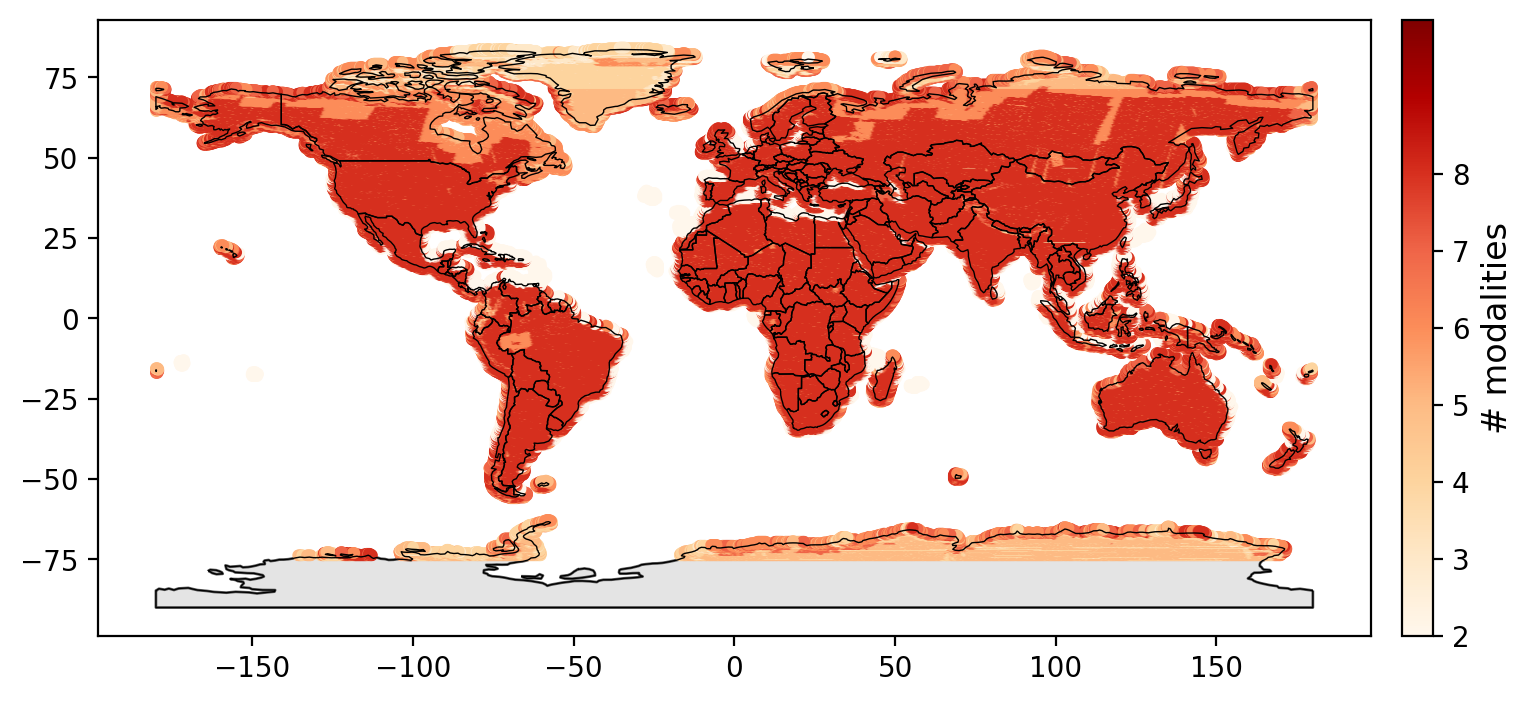}
    \caption{Global distribution of the Copernicus-Pretrain dataset.}
    \label{fig:datasetdistribution}
\end{figure}

In total, Copernicus-Pretrain consists of $\sim$18M images grouped into $\sim$310K $0.25\degree \times 0.25\degree$ cells, densely covering the whole land surface and near-land ocean with eight distinct Sentinel modalities. On average, each cell contains 4 local S1/S2 time series pairs (16 images each), 8 S3 images, 7/6/5/8 S5P CO/NO\textsubscript{2}/SO\textsubscript{2}/O\textsubscript{3} images, and 1 DEM image.

\cref{fig:datasetdistribution} shows the global distribution of the full dataset, while the detailed statistics are shown in \cref{tab:datasetstat}. The distribution and statistics of the all-modality-aligned subset and other detailed analyses can be found in the appendix \ref{app:ssl4eos}.
\section{Copernicus-FM}
\label{sec:dofas}

\begin{figure*}[ht]
    \centering
    \includegraphics[width=0.9\linewidth]{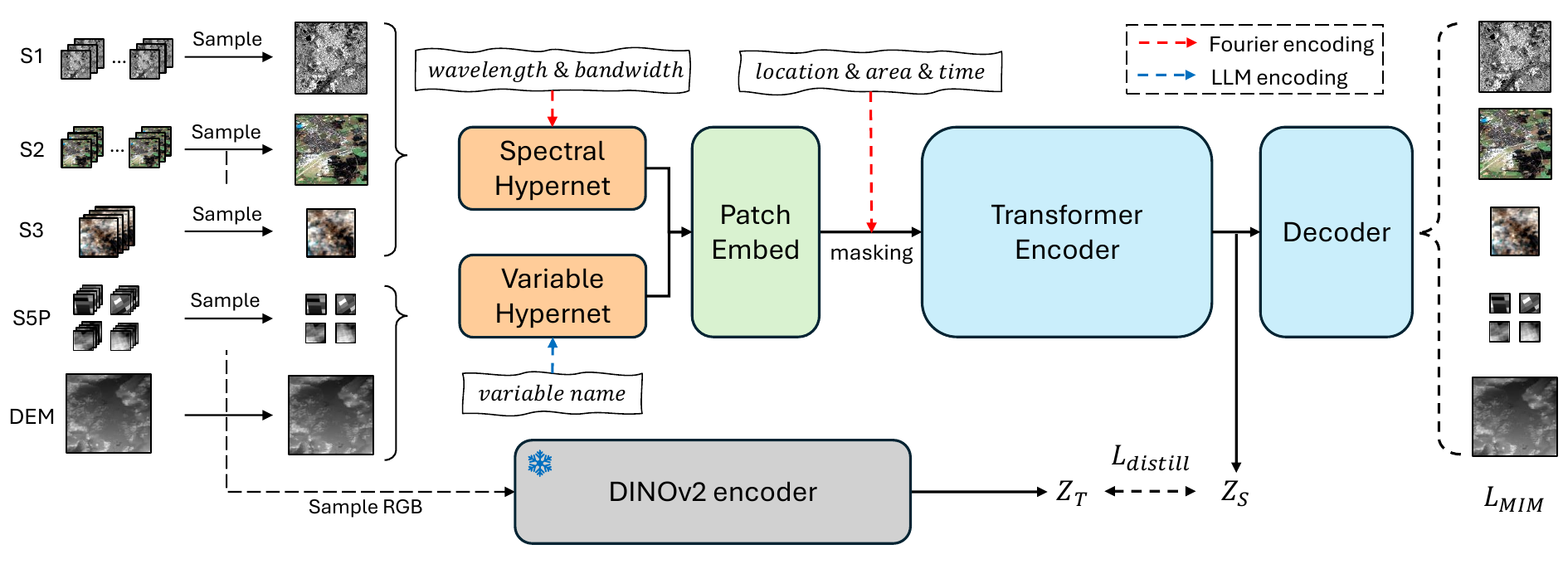}
    \caption{The general pretraining pipeline of Copernicus-FM. One image for each modality is sampled from a common grid cell in Copernicus-Pretrain, which is then patchified with kernel weights generated by the spectral or variable hypernetwork, based on the modality's spectral response or variable name. Further, Fourier-encoded metadata encodings are incorporated into the patch tokens. We conduct masked image modeling with auxiliary continual distillation for pretraining: masking and reconstructing masked-out patches for each modality, and distilling S1/2 or S2-derived RGB representations from powerful specialized teachers such as DINOv2~\cite{oquab2023dinov2}. }
    \label{fig:model_main}
\end{figure*}

% \begin{figure}[]
%     \centering
%     \includegraphics[width=0.9\linewidth]{figures/model_hypernet_v3.pdf}
%     \caption{The spectral and variable hypernetworks of Copernicus-FM.}
%     \label{fig:model_hypernet}
% \end{figure}

% % \begin{figure*}[]
% %     \centering
% %     \includegraphics[width=0.9\linewidth]{figures/model_meta.pdf}
% %     \caption{The metadata encoding of Copernicus-FM.}
% %     \label{fig:model_meta}
% % \end{figure*}

% \begin{figure}[]
%     \centering
%     \includegraphics[width=\linewidth]{figures/model_meta_v2.pdf}
%     \caption{The metadata encoding of Copernicus-FM.}
%     \label{fig:model_meta}
% \end{figure}

Based on the Copernicus-Pretrain dataset, we introduce Copernicus-FM, a new EO foundation model that 1) can process any spectral or non-spectral input modalities with varied spatial resolutions, and 2) when available, flexibly integrates metadata information like geolocation, geographical area, and time. \cref{fig:model_main} illustrates the general pretraining framework of Copernicus-FM, which uses unified hypernetworks to dynamically patchify different modalities into patch tokens, integrates metadata as Fourier encodings added to the patch tokens, and performs masked image modeling (MIM) with auxiliary continual distillation as the training objectives.

\subsection{Model architecture}

Given a grid sample unit with all modalities, one image is sampled from the local patches (for S1/S2) and time series for each modality. This results in 8 images with varied spatial and channel sizes from distinct modalities, serving as the main input to the model.

\vspace{-0.5em}
\paragraph{Dynamic patch embedding via sensor-aware hypernetworks}

We use a single unified architecture to deal with different input modalities. This is implemented by extending the wavelength-conditioned dynamic patch embedding design from DOFA~\cite{xiong2024neural}, where the core idea is to use a hypernetwork to dynamically generate kernel weights for the 2D convolution patch embedding layer.

As presented in \cref{fig:model_main}, denoting one spectral image (e.g., from S1/2/3) as $\mathbf{X} \in \mathbb{R}^{C \times H \times W}$, each channel has a corresponding central wavelength $\lambda \in \{{\lambda}_{1}, ..., {\lambda}_{C}\}$ and bandwidth $\delta \in \{{\delta}_{1}, ..., {\delta}_{C}\}$ acquired from the sensor's spectral response. The wavelengths and bandwidths serve as input to the spectral hypernetwork to generate patch embedding weights. 

First, we use a Fourier encoding as introduced in \citet{bodnar2024aurora} to encode a wavelength or bandwidth value into a $D$-dimensional vector, 
\begin{equation}
\label{eq:fourier}
\operatorname{FE}(x)=\left[\cos \frac{2 \pi x}{\omega_i}, \sin \frac{2 \pi x}{\omega_i}\right],\quad \forall \; 0 \leq i<D / 2,
\end{equation}
where $\omega_i$ are log-spaced values between the minimum and maximum: 
\begin{equation}
\label{eq:omega}
\omega_i=\exp \left(\log \omega_{\min }+i \cdot \frac{\log \omega_{\max }-\log \omega_{\min }}{D / 2-1}\right).
\end{equation}
One specific Fourier encoding is designed for wavelengths and bandwidths respectively, where $\omega_{\min}$ and $\omega_{\max}$ are defined based on the corresponding value ranges. The resulting vectors $\mathbf{V}_\lambda \in \mathbb{R}^{C \times D}$ and $\mathbf{V}_\delta \in \mathbb{R}^{C \times D}$ are then added together to form the spectral encodings $\mathbf{V}_\text{spec} \in \mathbb{R}^{C \times D}$. 

Following DOFA~\cite{xiong2024neural}, $\mathbf{V}_\text{spec}$ are further
transformed through MLP and multi-head attention layers to get weight vectors $\mathbf{M}_{\mathbf{w}} \in \mathbb{R}^{C \times p^2 D}$ and bias vectors $\mathbf{M}_{\mathbf{b}} \in \mathbb{R}^{C \times D}$, where $p$ is the expected patch size. The weight vectors are then reshaped into the convolution kernel $\mathbf{K}_{\mathrm{conv}} \in \mathbb{R}^{D \times C \times p \times p}$. Originally, a fixed patch size is required for all modalities, which can be acceptable when the scales do not differ much. However, the significant resolution gaps between our input modalities will lead to exploded memory and compute costs. To tackle this issue, we adapt the idea of FlexiVit~\cite{beyer2023flexivit} to dynamically reshape the kernel weights into a suitable patch size for each modality. Finally, the convolution operation for patch embedding is performed using the reshaped weights $\mathbf{K'}_{\mathrm{conv}}$ and biases $\mathbf{M}_{\mathbf{b}}$.

% \begin{equation}
% \label{eq:patchembed}
% \text { PatchEmbedding }:=\operatorname{Conv}\left(\mathbf{X}, \mathbf{K}_{\text {conv }}, \mathbf{M}_{\mathbf{b}}\right)
% \end{equation}

The above spectral hypernetwork can process any modality with a spectral response regardless of spatial and channel dimensions. However, it can not natively process non-spectral modalities such as atmospheric constituents from S5P and elevation maps from DEM. To bridge this gap, we introduce a variable hypernetwork to generate weights for non-spectral modalities. Since these modalities do not have a common meta-attribute like wavelength or bandwidth, we propose to directly encode their variable names using modern large language models (LLMs) with general scientific knowledge. Specifically, we use one frozen LLM encoder to encode the variable names of non-spectral inputs into $D$-dimensional vectors. This is done as a preprocessing step through one-time inference offline, thereby introducing zero additional cost to the general model framework. Similar to the spectral hypernetwork, the language-guided variable encodings $\mathbf{V}_{var}$ are further
transformed through MLP and attention layers to get weight and bias vectors, and reshaped to modality-specific patch sizes for the convolution kernel of the patch embedding layer.

\vspace{-0.5em}
\paragraph{Metadata integration with unified Fourier encoding}

The dynamic patch embedding layer patchifies the input image into a sequence of patch tokens. In the standard Transformer architecture, positional encodings $\mathbf{P} \in \mathbb{R}^{N \times D}$ (where $N$ is the number of patches) are added along with a classification token to the patch tokens. In Copernicus-FM, we introduce additional metadata encodings in parallel to the positional encodings to integrate metadata information when available. Similar to wavelength and bandwidth encoding in the spectral hypernetwork, we again use a Fourier encoding (\cref{eq:fourier,eq:omega}) to unify different metadata with their corresponding value ranges.

\begin{figure*}[htbp]
    \centering
    \includegraphics[width=0.9\linewidth]{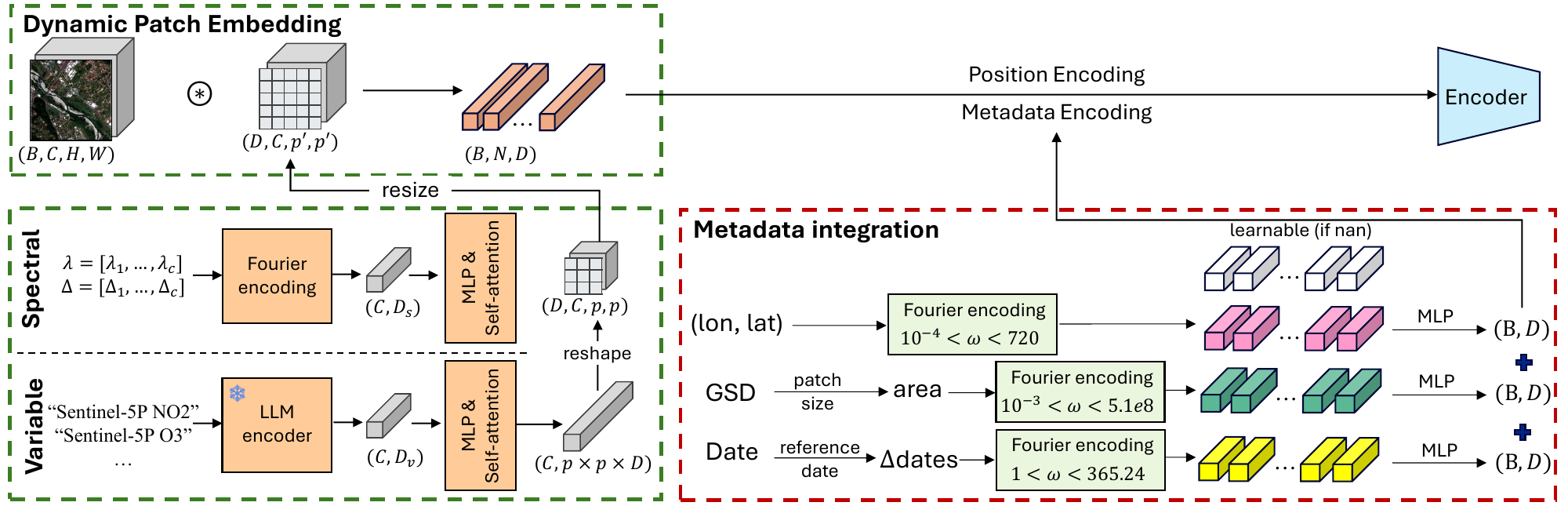}
    \caption{Dynamic patch embedding (left) and metadata integration (right) of Copernicus-FM.}
    \label{fig:model_encoding}
\end{figure*}

As shown in \cref{fig:model_encoding}, we consider three common types of metadata for one input image: geolocation, spatial coverage, and time. For geolocation, the central coordinates of one image (longitude and latitude) are encoded into $D/2$-dimensional vectors and then concatenated to the location encoding $\text{Loc} \in \mathbb{R}^{D}$. For spatial coverage, the patch area (in km\textsuperscript{2}) is calculated from the ground sample distance (GSD) and patch size, and then encoded into the area encoding $\text{Area} \in \mathbb{R}^{D}$. For time, the temporal difference (in days) between the acquisition date of the image and a reference date is calculated, and then encoded into the time encoding $\text{Time} \in \mathbb{R}^{D}$. These metadata encodings are further processed by a corresponding MLP layer, expanded across the patch dimensions, and added together into the positional encodings. In practice, the metadata may not always be available. Therefore, we introduce a learnable token for each metadata when it is unavailable. We simulate such metadata-missing scenarios by randomly dropping part of the metadata during each iteration.

\begin{table*}[htbp]
\centering
\caption{Ablation study of  Copernicus-FM. OA: overall accuracy, mAP: mean average precision, and RMSE: root mean squared error.}
% \fontsize{9.5pt}{9.5pt}\selectfont
\small
\label{tab:ablation-main}
    \begin{tabular}{llllll}
        \toprule
        \multirow{2}{*}{} & EuroSAT-S1 & EuroSAT-S2 & EuroSAT-RGB & LC100-S3 & AQ-O3-S5P \\
        & \multicolumn{3}{c}{(OA $\uparrow$)} & (mAP $\uparrow$) & (RMSE $\downarrow$) \\ \midrule
        Baseline~\cite{xiong2024neural} + dynamic patch size & 56.3 & 87.6 & 62.2 & 86.7 & 2218.0 \\
        ... + bandwidth (Fourier encoding) & 56.5 \inc{0.2} & 88.9 \inc{1.3} & 65.4 \inc{3.2} & 87.1 \inc{0.4} & 1710.7 \decreg{507.3} \\
        ... + variable hypernetwork & 57.5 \inc{1.0} & 88.9 \inc{0.0} & 65.8 \inc{0.4} & 86.6 \dec{0.5} & 1598.1 \decreg{112.6} \\
        ... + metadata encoding & 77.9 {\bf \inc{22.4}} & 88.9 \inc{0.0} & 78.5 {\bf \inc{12.7}} & 90.7 {\bf \inc{4.1}} & 839.3 {\bf \decreg{758.8}} \\
        ... + continual distillation & 81.0 \inc{2.9} & 89.5 {\bf \inc{0.6}} & 78.9 \inc{0.4} & 90.7 \inc{0.0} & 811.6 \decreg{27.7} \\ \bottomrule
    \end{tabular}
\end{table*}

\subsection{Training objectives}

Following DOFA~\cite{xiong2024neural}, we combine MIM and continual distillation for the pretraining of Copernicus-FM. Specifically, MIM is conducted by MAE-style~\cite{he2022masked} masked reconstruction for each modality---the patch tokens are randomly masked out and sent through the encoder and a lightweight Transformer decoder to reconstruct the masked patches. A dynamic patch predictor similar to the patch embedding layer in the encoder is used for prediction. Following \citet{he2022masked}, we conduct reconstruction directly on the flattened patch tokens. Thus, a fully-connected layer is used at the end of the predictor instead of a convolutional layer.

Meanwhile, we conduct auxiliary continual distillation with a small loss weight, using powerful single-modal or general-domain foundation models to serve as an anchor to guide and refine the latent space. For example, we distill S2-derived RGB representations from a frozen DINOv2~\cite{oquab2023dinov2} encoder with a cosine-similarity loss. This helps improve the out-of-the-box representation quality of MIM-based pretraining, reduces compute cost with faster convergence, and also implicitly boosts the model's general knowledge beyond our specific pretraining data sources.

\subsection{Implementation details}

We pretrain Copernicus-FM with ViT-Base on the Copernicus-Pretrain dataset (220K grids with all modalities available) for 100 epochs. Data augmentations include simple resized cropping and horizontal flipping.
Input image sizes are as shown in \cref{fig:model_main}, with patch sizes $16 \times 16$ for S1/2, $8\times8$ for S3, $4\times4$ for S5P, and $64\times64$ for DEM. We use Llama~3.2~\cite{dubey2024llama} to encode variable names. The Fourier encoding value ranges are [1e2, 1e9] for wavelengths, [1, 1e9] for bandwidths, [1e-4, 720] for longitudes/latitudes, [1e-3, 5.1e8] for patch area, and [1, 365.25] for time. The drop probability for metadata is 0.7. The masking ratio for MIM is 0.7. We distill RGB from DINOv2~\cite{oquab2023dinov2}, and S1/2 from SoftCon~\cite{wang2024multi} with loss weights of 0.1 and 0.2, respectively. More details can be found in the appendix \ref{app:Copernicus-FM}.

\subsection{Ablation studies}

We conduct ablation studies on different components of Copernicus-FM with ViT-Small on a 10K subset of Copernicus-Pretrain, and evaluate the pretrained encoders on a range of downstream tasks covering different modalities. Specifically, we conduct $k$-NN classification on EuroSAT-SAR (S1)~\cite{wang2024feature}, EuroSAT-MS (S2)~\cite{helber2019eurosat}, and EuroSAT-RGB~\cite{helber2019eurosat}, linear probing on LC100Cls-S3 (multi-label), and dense regression with frozen encoder on AQ-O3-S5P (air pollutant regression of O\textsubscript{3}). %The latter two datasets will be further described in the next section.

The main ablation results are shown in \cref{tab:ablation-main}, where consistent improvement can be observed when gradually adding the spectral hypernetwork bandwidth, variable hypernetwork, metadata encoding, and continual distillation. Among them, the benefits of metadata encoding appear to be the most significant, especially in non-optical modalities. This highlights the importance of metadata beyond pure imagery for remote sensing applications. Subtractive ablation, ablation on each metadata (location matters), the metadata dropping ratio (higher is better), as well as other minor ablations can be found in the appendix \ref{app:Copernicus-FM}.

%\newpage
\section{Copernicus-Bench}
\label{sec:sentinelbench}

\begin{table*}[htbp]
    \centering
    \caption{Characteristics of datasets in Copernicus-Bench. seg, cls, cd, and reg represent segmentation, classification, change detection, and regression, respectively. *Time series support (default mode is 1 image for seg and 2 for cd). $^\dagger$Geolocation metadata not available.}
    \label{tab:senbench}
    %\small
    \begin{adjustbox}{max width=\textwidth}
    \begin{tabular}{llcccccc}
        \toprule
        \textbf{Level} & \textbf{Name} & \textbf{Task} & \textbf{\# Images} & \textbf{Image Size} & \textbf{\# Classes} & \textbf{Source} & \textbf{License} \\
        \midrule
        % \multicolumn{8}{c}{\textbf{Level-1 (Preprocessing)}} \\
        % \midrule
        \multirow{2}{*}{L1} & Cloud-S2 & seg & 1699/567/551 & 512$\times$512$\times$13 & 4 & CloudSEN12+~\cite{aybar2024cloudsen12+} & CC0-1.0 \\
         & Cloud-S3 & seg & 1197/399/399 & 256$\times$256$\times$21 & 5 & new & CC-BY-4.0 \\
        \midrule
        % \multicolumn{8}{c}{\textbf{Level-2 (Base Applications)}} \\
        % \midrule
        \multirow{8}{*}{L2} & EuroSAT-S1 & cls & 16200/5400/5400 & 64$\times$64$\times$2 & 10 & EuroSAT-SAR~\cite{wang2024feature} & CC-BY-4.0 \\
         & EuroSAT-S2 & cls & 16200/5400/5400 & 64$\times$64$\times$13 & 10 & EuroSAT~\cite{helber2019eurosat} & MIT \\
         & BigEarthNet-S1 & cls & 11894/6117/5991 & 120$\times$120$\times$2 & 19 & BigEarthNet v2.0~\cite{clasen2024reben} & CDLA-Permissive-1.0 \\
         & BigEarthNet-S2 & cls & 11894/6117/5991 & 120$\times$120$\times$12 & 19 & BigEarthNet v2.0~\cite{clasen2024reben} & CDLA-Permissive-1.0 \\
         & LC100Cls-S3 & cls & 5181/1727/1727* & 96$\times$96$\times$21 & 23 & new & CC-BY-4.0 \\
         & DFC2020-S1$^\dagger$ & seg & 3156/986/986 & 256$\times$256$\times$2 & 10 & DFC2020~\cite{rha7-m332-19} & CC-BY-4.0 \\
         & DFC2020-S2$^\dagger$ & seg & 3156/986/986 & 256$\times$256$\times$13 & 10 & DFC2020~\cite{rha7-m332-19} & CC-BY-4.0 \\
         & LC100Seg-S3 & seg & 5181/1727/1727* & 96$\times$96$\times$21 (288$\times$288) & 23 & new & CC-BY-4.0 \\
        \midrule
        % \multicolumn{8}{c}{\textbf{Level-3 (Specialized Applications)}} \\
        % \midrule
        \multirow{5}{*}{L3} & Flood-S1 & cd & 3000/1000/1000* & 224$\times$224$\times$2 & 3 & Kuro Siwo~\cite{bountos2023kuro} & MIT \\
         & LCZ-S2$^\dagger$ & cls & 15000/5000/5000 & 32$\times$32$\times$10 & 17 & So2Sat LCZ42~\cite{zhu2020so2sat} & CC-BY-4.0 \\
         & Biomass-S3 & reg & 3000/1000/1000* & 96$\times$96$\times$21 (288$\times$288) & - & new & CC-BY-4.0 \\
         & AQ-NO2-S5P & reg & 1480/493/494* & 56$\times$56$\times$1 & - & new & CC-BY-4.0 \\
         & AQ-O3-S5P & reg & 1480/493/494* & 56$\times$56$\times$1 & - & new & CC-BY-4.0 \\
        \bottomrule
    \end{tabular}
    \end{adjustbox}

\end{table*}

\begin{table*}[ht]
\centering
\caption{Benchmark results with representative single-, dual-, and multi-modal foundation models on Copernicus-Bench. We report three-run averages with standard deviations. *: patch size 16 for S1/2 (8 for S3, 4 for S5P). Random: encoder random init. Best scores in \textbf{bold}.}
\label{tab:benchmark}
%\small
\renewcommand{\arraystretch}{1.2} % Adjust row spacing
\begin{adjustbox}{max width=\textwidth}
\begin{tabular}{lc
>{\columncolor[HTML]{EFEFEF}}c 
>{\columncolor[HTML]{EFEFEF}}c ccccc}
\toprule 
 & \textbf{Metric} & \textbf{Supervised} & \textbf{Supervised} & \textbf{Random} & \textbf{SoftCon~\cite{wang2024multi}} & \textbf{CROMA~\cite{fuller2024croma}} & \textbf{DOFA~\cite{xiong2024neural}} & \textbf{Copernicus-FM} \\ \midrule
Backbone & -- & ViT-S/16 & ViT-B/16 & ViT-B/16 & ViT-B/14 & ViT-B/8 & ViT-B/16 & ViT-B/16* \\
Modality & -- & -- & -- & -- & S1/S2 & S1+S2 & All (spectral) & All \\ \midrule
Cloud-S2 & mIoU & 64.2 $\pm$ 0.9 & 59.4 $\pm$ 1.0 & 60.4 $\pm$ 0.2 & 66.9 $\pm$ 0.3 & 65.0 $\pm$ 0.2 & 65.0 $\pm$ 0.2 & \textbf{66.7 $\pm$ 0.1} \\
Cloud-S3 & mIoU & 61.7 $\pm$ 0.7 & \textbf{63.0 $\pm$ 0.8} & 60.9 $\pm$ 0.0 & -- & -- & 58.2 $\pm$ 0.1 & 62.0 $\pm$ 0.7 \\
EuroSAT-S1 & OA & 81.7 $\pm$ 0.7 & 81.5 $\pm$ 0.9 & 75.4 $\pm$ 0.4 & 83.6 $\pm$ 0.1 & 83.9 $\pm$ 0.1 & 81.7 $\pm$ 0.1 & \textbf{87.2 $\pm$ 0.1} \\
EuroSAT-S2 & OA & 97.5 $\pm$ 0.0 & 97.6 $\pm$ 0.1 & 92.5 $\pm$ 0.1 & 96.7 $\pm$ 0.0 & 97.0 $\pm$ 0.1 & 97.2 $\pm$ 0.1 & \textbf{97.9 $\pm$ 0.1} \\
BigEarthNet-S1 & mAP & 71.5 $\pm$ 0.4 & 70.6 $\pm$ 0.4 & 63.8 $\pm$ 0.1 & \textbf{78.7 $\pm$ 0.0} & 70.8 $\pm$ 0.0 & 70.5 $\pm$ 0.0 & 77.9 $\pm$ 0.0 \\
BigEarthNet-S2 & mAP & 79.5 $\pm$ 0.5 & 80.1 $\pm$ 0.1 & 71.6 $\pm$ 0.1 & \textbf{83.6 $\pm$ 0.0} & 76.4 $\pm$ 0.0 & 75.5 $\pm$ 0.0 & 79.0 $\pm$ 0.0 \\
LC100Cls-S3 & mAP & 91.3 $\pm$ 0.3 & 91.4 $\pm$ 0.5 & 88.9 $\pm$ 0.1 & -- & -- & 89.5 $\pm$ 0.0 & \textbf{93.3 $\pm$ 0.4} \\
DFC2020-S1 & mIoU & 49.9 $\pm$ 0.4 & 50.8 $\pm$ 0.5 & 45.4 $\pm$ 0.1 & \textbf{52.8 $\pm$ 0.6} & \textbf{52.7 $\pm$ 0.1} & 49.7 $\pm$ 0.1 & 52.4 $\pm$ 0.1 \\
DFC2020-S2 & mIoU & 65.3 $\pm$ 0.6 & 66.2 $\pm$ 0.7 & 62.3 $\pm$ 0.0 & 64.1 $\pm$ 0.3 & \textbf{66.5 $\pm$ 0.0} & 61.8 $\pm$ 0.1 & 64.5 $\pm$ 0.1 \\
LC100Seg-S3 & mIoU & 20.1 $\pm$ 0.4 & 19.3 $\pm$ 0.5 & 18.2 $\pm$ 0.1 & -- & -- & 16.5 $\pm$ 0.1 & \textbf{24.1 $\pm$ 0.0} \\
Flood-S1 & mIoU & 78.0 $\pm$ 0.1 & \textbf{78.3 $\pm$ 0.3} & 75.1 $\pm$ 0.1 & 77.2 $\pm$ 0.1 & 77.4 $\pm$ 0.1 & 76.0 $\pm$ 0.1 & 77.7 $\pm$ 0.0 \\
LCZ-S2 & OA & \textbf{86.6 $\pm$ 0.7} & 85.3 $\pm$ 0.8 & 77.4 $\pm$ 0.1 & 83.6 $\pm$ 0.2 & 84.1 $\pm$ 0.0 & 83.0 $\pm$ 0.3 & 84.4 $\pm$ 0.0 \\
Biomass-S3 & RMSE $\downarrow$ & 68.1 $\pm$ 0.3 & 68.3 $\pm$ 0.4 & 68.7 $\pm$ 0.5 & -- & -- & 74.1 $\pm$ 0.1 & \textbf{66.3 $\pm$ 0.1} \\
AQ-NO2-S5P & RMSE $\downarrow$ & 3.4 $\pm$ 0.0 & 3.4 $\pm$ 0.0 & 3.4 $\pm$ 0.0 & -- & -- & 3.3 $\pm$ 0.0 & \textbf{2.8 $\pm$ 0.0} \\
AQ-O3-S5P & RMSE $\downarrow$ & 1781.3 $\pm$ 29.8 & 1766.8 $\pm$ 22.1 & 1741.6 $\pm$ 11.5 & -- & -- & 1755.6 $\pm$ 19.8 & \textbf{789.4 $\pm$ 2.6} \\ \bottomrule
\end{tabular}
\end{adjustbox}
\end{table*}

For a thorough evaluation of Copernicus-FM, we curate a benchmark suite, Copernicus-Bench, covering various Sentinel missions with hierarchical downstream tasks. 

\subsection{Datasets in Copernicus-Bench}

Copernicus-Bench consists of 15 datasets with varied Sentinel modalities organized into three application levels in practice: preprocessing, base applications, and specialized applications. Level-1 includes two cloud detection tasks from S2/3, level-2 includes 8 land use land cover classification/segmentation tasks from S1/2/3, and level-3 includes 5 specialized tasks from S1/2/3/5P. Among all the datasets, 9 are derived from existing datasets with permissive licenses, and 6 are newly curated to fill in the gap in ML-ready datasets for S3/S5P. The detailed characteristics are shown in \cref{tab:senbench}. The curation process for each dataset and detailed benchmark analyses can be found in the appendix \ref{app:cobench}.

\begin{table*}[ht]
\centering
\caption{Linear regression results on 10-year mean and standard deviation (variability) of 6 climate parameters. We report three-run average RMSE scores with standard deviation (error bar). Best scores in \textbf{bold}.}
\label{tab:climate}
\begin{adjustbox}{max width=\textwidth}
\begin{tabular}{lcccccccccccc}
\toprule
\multirow{2}{*}{Input Source} & 
\multicolumn{2}{c}{Temperature (°C)} & 
\multicolumn{2}{c}{Precipitation (m)} & 
\multicolumn{2}{c}{Surface press. (Pa)} & 
\multicolumn{2}{c}{Sea-level press. (Pa)} & 
\multicolumn{2}{c}{$u$ wind (m/s)} & 
\multicolumn{2}{c}{$v$ wind (m/s)} \\
\cmidrule(lr){2-3} \cmidrule(lr){4-5} \cmidrule(lr){6-7} \cmidrule(lr){8-9} \cmidrule(lr){10-11} \cmidrule(lr){12-13}
 & Avg & Std & Avg & Std & Avg & Std & Avg & Std & Avg & Std & Avg & Std \\ 
\midrule
{Coord. (raw)} & 8.36 & 3.06 & 0.79 & 0.50 & 7305.91 & 141.38 & 290.68 & 164.84 & 1.12 & 0.55 & 0.95 & 0.58 \\
 & $\pm\,0.00$ & $\pm\,0.00$ & $\pm\,0.00$ & $\pm\,0.00$ & $\pm\,0.37$ & $\pm\,0.01$ & $\pm\,0.02$ & $\pm\,0.01$ & $\pm\,0.00$ & $\pm\,0.00$ & $\pm\,0.00$ & $\pm\,0.00$ \\ 

{Coord. (FE)} & 3.99 & 2.09 & 0.61 & 0.42 & 6537.54 & 101.09 & 200.93 & 108.12 & 0.96 & 0.51 & 0.80 & 0.54 \\
 & $\pm\,0.00$ & $\pm\,0.00$ & $\pm\,0.00$ & $\pm\,0.00$ & $\pm\,1.78$ & $\pm\,0.18$ & $\pm\,0.09$ & $\pm\,0.01$ & $\pm\,0.00$ & $\pm\,0.00$ & $\pm\,0.00$ & $\pm\,0.00$ \\ 

{Embed.} & 2.66 & 1.72 & 0.47 & 0.33 & \textbf{1627.21} & 97.53 & 197.17 & 117.57 & 0.94 & 0.43 & 0.78 & \textbf{0.43} \\ 
 & $\pm\,0.01$ & $\pm\,0.00$ & $\pm\,0.00$ & $\pm\,0.00$ & $\pm\,30.13$ & $\pm\,0.71$ & $\pm\,0.62$ & $\pm\,0.86$ & $\pm\,0.00$ & $\pm\,0.00$ & $\pm\,0.00$ & $\pm\,0.00$ \\ 

{Embed. + coord.} & \textbf{1.98} & \textbf{1.35} & \textbf{0.46} & \textbf{0.32} & 1676.39 & \textbf{78.53} & \textbf{174.80} & \textbf{90.37} & \textbf{0.91} & \textbf{0.42} & \textbf{0.76} & \textbf{0.43} \\
 & $\pm\,0.01$ & $\pm\,0.00$ & $\pm\,0.00$ & $\pm\,0.00$ & $\pm\,33.64$ & $\pm\,0.08$ & $\pm\,0.21$ & $\pm\,0.15$ & $\pm\,0.00$ & $\pm\,0.00$ & $\pm\,0.00$ & $\pm\,0.00$ \\ 

\rowcolor[HTML]{EFEFEF} 
{Coord. (SatCLIP~\cite{klemmer2023satclip})} & 3.67 & 1.58 & 0.48 & \textbf{0.32} & 3178.91 & 97.74 & 202.76 & 110.71 & 0.92 & 0.43 & 0.79 & 0.45 \\
\rowcolor[HTML]{EFEFEF} 
 & $\pm\,0.03$ & $\pm\,0.00$ & $\pm\,0.00$ & $\pm\,0.00$ & $\pm\,8.75$ & $\pm\,0.19$ & $\pm\,0.27$ & $\pm\,0.41$ & $\pm\,0.00$ & $\pm\,0.00$ & $\pm\,0.00$ & $\pm\,0.00$ \\ 
\bottomrule
\end{tabular}
\end{adjustbox}
\vspace{-0.5em}
\end{table*}

\subsection{Benchmark results}

We benchmark supervised training and frozen-encoder evaluation of several representative single-, dual-, and multimodal foundation models on Copernicus-Bench. For classification tasks, we perform linear probing with batch size 64 and the SGD optimizer for 50 epochs; for segmentation and regression tasks, we train a UPerNet~\cite{xiao2018unified} decoder on top of the frozen encoder with batch size 16 and the AdamW optimizer for 50 epochs; for change detection tasks, we separately encode the pre- and post-event images and calculate their feature map differences as input to a UPerNet decoder, hyperparameters following the segmentation design. For supervised baselines, we train the model from scratch for 80 epochs. We conduct a simple grid search for the learning rates and report test metrics under the best validation scores. All experiments are conducted on a single GPU with three runs to get the mean and standard deviation. The results are shown in \cref{tab:benchmark}. It can be seen that our Copernicus-FM is comparable to or better than state-of-the-art single- or multimodal foundation models on their applicable tasks, and largely improves downstream performances on S3 and S5P tasks. Encouragingly, our results outperform supervised training on the majority (11/15) of tasks, despite utilizing fewer trainable parameters and iteration steps. The results also verify the benefits of cross-modal pretraining on both surface and atmospheric applications.

\section{Bridging EO \& climate via grid embeddings}
\label{sec:climate}

While Copernicus-FM has demonstrated its benefits for various EO tasks, one of its most exciting potentials lies in bridging EO with weather and climate analysis---an intersection that is still largely unexplored. Thanks to the grid-based dataset structure in Copernicus-Pretrain, we achieve a direct alignment between EO imagery and climate parameters from ERA5\cite{hersbach2020era5}, one of the most widely used reanalysis datasets for weather and climate research. This allows Copernicus-FM-encoded grid embeddings to serve as semantically rich geographical representations that could potentially enhance climate modeling. 

To evaluate this potential, we curate a simple set of climate tasks predicting the 10-year mean and standard deviation of 6 important climate parameters (2~m mean air temperature, annual total precipitation, surface pressure, mean sea-level pressure, and $u$/$v$ component of 10-meter wind speed) using geocoordinates or geographical representations. We randomly sample 10K grids around the globe, split them into train/val/test subsets, and calculate the corresponding climate parameters from ERA5 as targets. As a baseline, the central coordinates of the grids are sent through a linear regression model for prediction. As shown in \cref{tab:climate}, the two baseline models with raw or Fourier-encoded coordinates as input verify the basic correlation between geography and climate. We then use Copernicus-FM-derived grid embeddings as input, obtained by simply averaging model-encoded images from various modalities. These embeddings provide a richer, high-level representation of each grid's environmental characteristics and consistently outperform coordinate-based baselines. Moreover, combining grid embeddings with coordinates yields the best overall performance, demonstrating the complementary nature of  spatial context and EO-derived semantic features. Notably, even when compared against location encodings from a specialized location encoder trained on EO images~\cite{klemmer2023satclip}, the grid embeddings still outperform, emphasizing the unique value of EO representations for climate studies.
We extend a global embedding dataset at 0.25$\degree$ based on these promising findings (\cref{fig:embed1}, see appendix \ref{app:climate} for details).
% Based on the promising findings above, we extend the grid embeddings to a global embedding dataset at 0.25$\degree$, Copernicus-Embed-025deg, from the whole Copernicus-Pretrain dataset (\cref{fig:embed1}, more details in the supplementary material).
Looking ahead, we envision incorporating these grid embeddings into ML-based training for medium-range weather forecasting, expanding the set of static variables with EO-derived visual representations, or dynamic variables with representation time series.

\begin{figure*}[ht]
    \centering
    \includegraphics[width=0.95\linewidth]{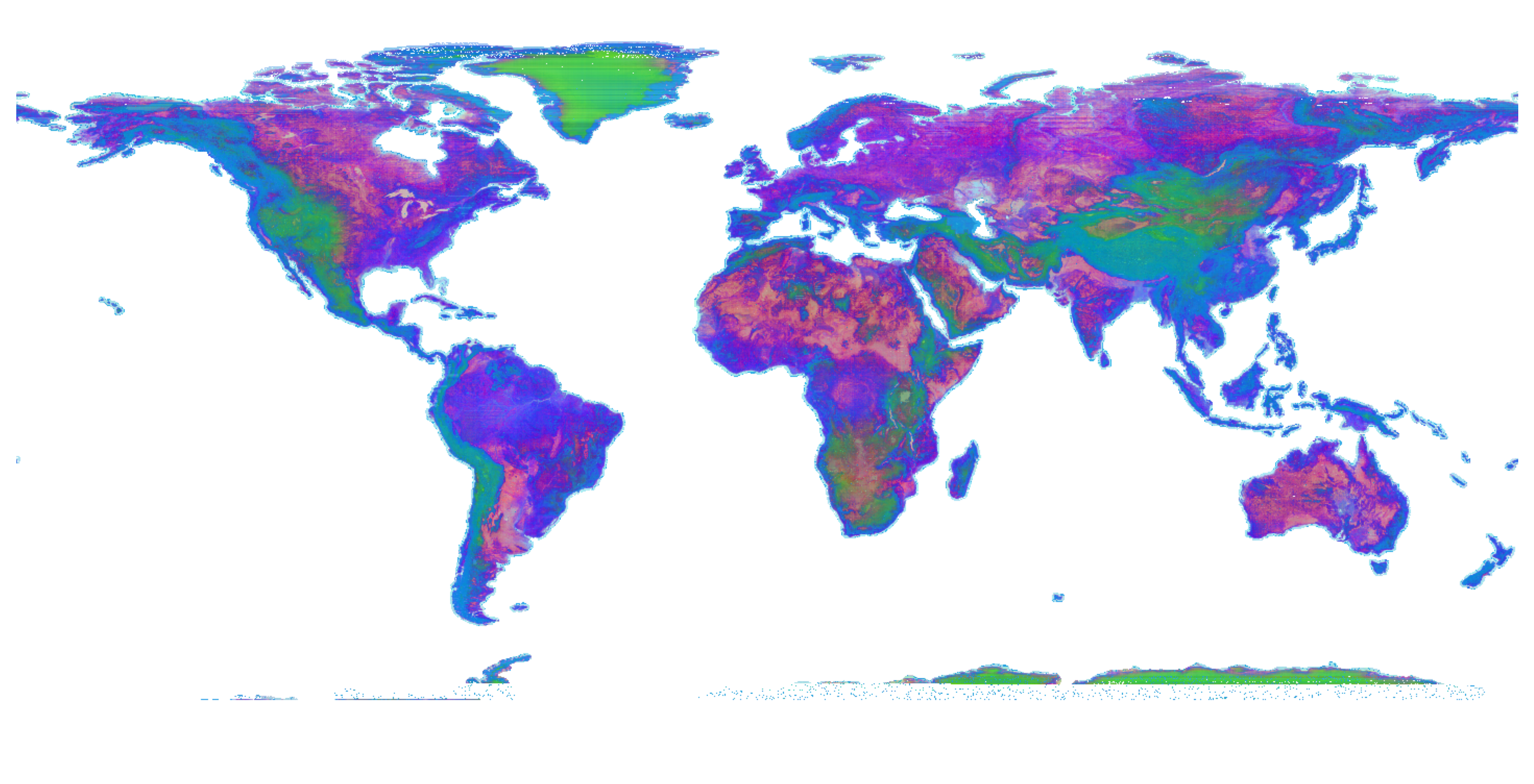}
    \caption{Visualization of the Copernicus-Embed-025deg dataset as a global embedding map (PCA to 3-dim).}
    \label{fig:embed1}
\end{figure*}

\section{Conclusion}
\label{sec:conclusion}

This work presents a series of efforts towards next-generation EO foundation models, advancing existing approaches in data, model, and benchmark. We introduce \textbf{Copernicus-Pretrain}, extending EO pretraining datasets to all major Sentinel missions, encompassing both Earth's surface and its atmosphere. We propose \textbf{Copernicus-FM}, leveraging dynamic hypernetworks to process any spectral or non-spectral modality, enhancing adaptability across diverse EO sensors. We establish \textbf{Copernicus-Bench}, a comprehensive benchmark with hierarchical downstream tasks across various Sentinel modalities. 
%In addition, we showcase the strong potential of our EO grid embeddings in a simple climate prediction task. This lays the foundation for future research in bridging EO with weather and climate studies—exploring the advantages of EO representations for both long- and short-term forecasting, and vice versa, towards the unified EO and climate foundation models.
Furthermore, we demonstrate the benefits of EO grid embeddings in a simple climate prediction task, highlighting their potential in bridging EO with weather and climate studies. This paves the way towards unified EO and climate foundation models, unlocking new possibilities for integrated Earth system understanding in future work. 

As a limitation, the scope of this work is restricted to the Sentinel series and a temporal range of about 1 year. Future work will thus also explore efficient scaling for satellite expansion, as well as native model support for multimodal fusion and time series processing.

%\clearpage

% \input{sec/2_formatting}
% \input{sec/3_finalcopy}
{
    \small
    \bibliographystyle{ieeenat_fullname}
    \bibliography{main}
}

\clearpage

\appendix
\onecolumn
\begin{center}
  {\LARGE \bfseries Supplementary Material}
\end{center}

\section{Copernicus-Pretrain}
\label{app:ssl4eos}

This section reports more detailed characteristics and statistical analyses for the Copernicus-Pretrain dataset.

\subsection{Comparison to existing EO pretraining datasets}
\cref{tab:eodatasets} shows a detailed comparison between Copernicus-Pretrain and several existing EO pretraining datasets.

\begin{table*}[ht]
\centering
\caption{A comparison of existing EO pretraining datasets.}
\label{tab:eodatasets}
\begin{tabular}{lcrrrr}
\toprule
Dataset & Modality & Resolution & \# Time stamps & \# patches & \# pixels \\ \midrule
fMoW~\cite{christie2018functional} & RGB, MS & 0.3--10~m & 3 & 2M & 50B \\
SEN12MS~\cite{schmitt2019sen12ms} & SAR, MS & 10~m & 1 & 540K & 35B \\
SeCo~\cite{manas2021seasonal} & MS & 10~m & 5 & 1M & 70B \\
SSL4EO-S12~\cite{wang2023ssl4eo} & SAR, MS & 10~m & 4 & 3M & 140B \\
SSL4EO-L~\cite{stewart2023ssl4eo} & MS & 30~m & 4 & 5M & 348B \\
SatlasPretrain~\cite{bastani2023satlaspretrain} & SAR, MS, RGB & 0.5--10~m & $\sim$10 & $>$10M & 17T \\
MMEarth~\cite{nedungadi2024mmearth} & SAR, MS, height, landcover, etc. & 10--15~m & 1 & 6M & 120B \\
SpectralEarth~\cite{braham2024spectralearth} & HS & 30~m & 1--23 & 540K & 10B \\
Major TOM~\cite{francis2024major} & SAR, MS & 10~m & 1 & 8M & 6.8T \\
Copernicus-Pretrain & SAR, MS, S3, DEM, S5P & 10~m--1~km & 1--12 & 19M & 920B \\ \bottomrule
\end{tabular}
\end{table*}

\subsection{Extended statistics}

\begin{figure}[ht]
    \centering
    \includegraphics[width=0.8\linewidth]{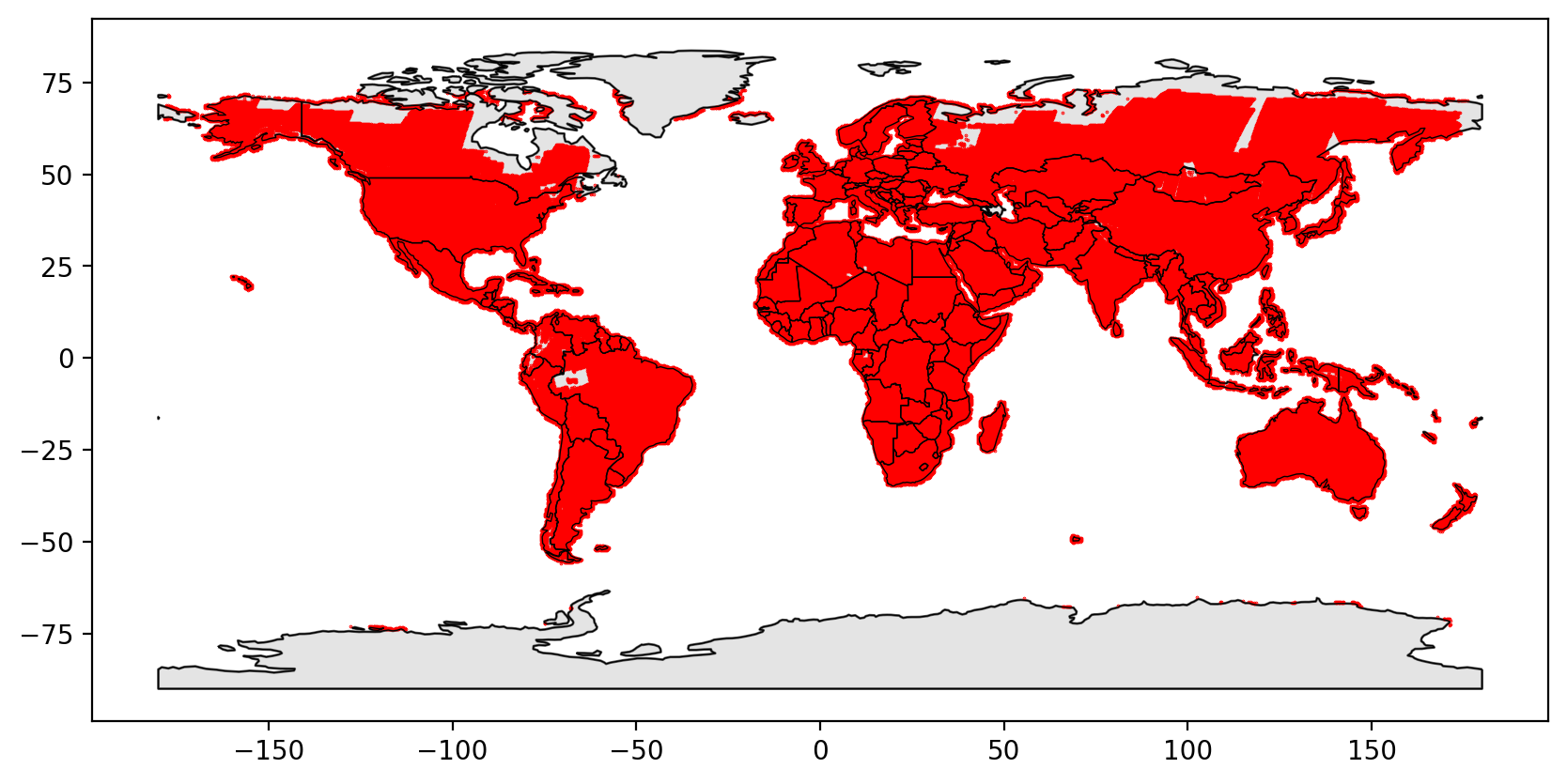}
    \caption{Global distribution of the joint subset of the Copernicus-Pretrain dataset.}
    \label{fig:datasetdistribution2}
\end{figure}

\paragraph{All-modality-aligned subset}
The Copernicus-Pretrain dataset contains 310K grids with at least one modality, of which 220K have all eight modalities. \cref{tab:datasetstat2} shows the detailed characteristics of the 220K subset, and \cref{fig:datasetdistribution2} presents its global distribution. We refer to the full dataset (grids with at least one modality) as ``union'', and the all-modality-aligned subset (grids with all modalities) as ``joint''.

\begin{table*}[ht]
\centering
\caption{Copernicus-Pretrain dataset characteristics (joint 220K subset).}
\label{tab:datasetstat2}
\begin{tabular}{lccccc}
\toprule
 & image size & \# grid cells & \# patches & \# timestamps & \# total images \\ \midrule
Sentinel-1 GRD & $\sim$264x264 & 219,543 & 996,978 & $\sim$4 & 3,948,217 \\
Sentinel-2 TOA & $\sim$264x264 & 219,543 & 996,978 & $\sim$4 & 3,948,217 \\
Sentinel-3 OLCI & $\sim$96x96 & 219,543 & 219,543 & $\sim$8 & 1,720,881 \\
Sentinel-5P CO & $\sim$28x28 & 219,543 & 219,543 & 1--12 & 1,548,349 \\
Sentinel-5P NO2 & $\sim$28x28 & 219,543 & 219,543 & 1--12 & 1,394,800 \\
Sentinel-5P SO2 & $\sim$28x28 & 219,543 & 219,543 & 1--12 & 1,188,864 \\
Sentinel-5P O3 & $\sim$28x28 & 219,543 & 219,543 & 1--12 & 1,750,542 \\
Copernicus DEM & $\sim$960x960 & 219,543 & 219,543 & 1 & 219,543 \\ \midrule
Copernicus-Pretrain & - & 219,543 & 3,311,214 & - & 15,720,353 \\ \bottomrule
\end{tabular}
\end{table*}

\begin{figure*}[ht]
    \centering
    \includegraphics[width=0.8\linewidth]{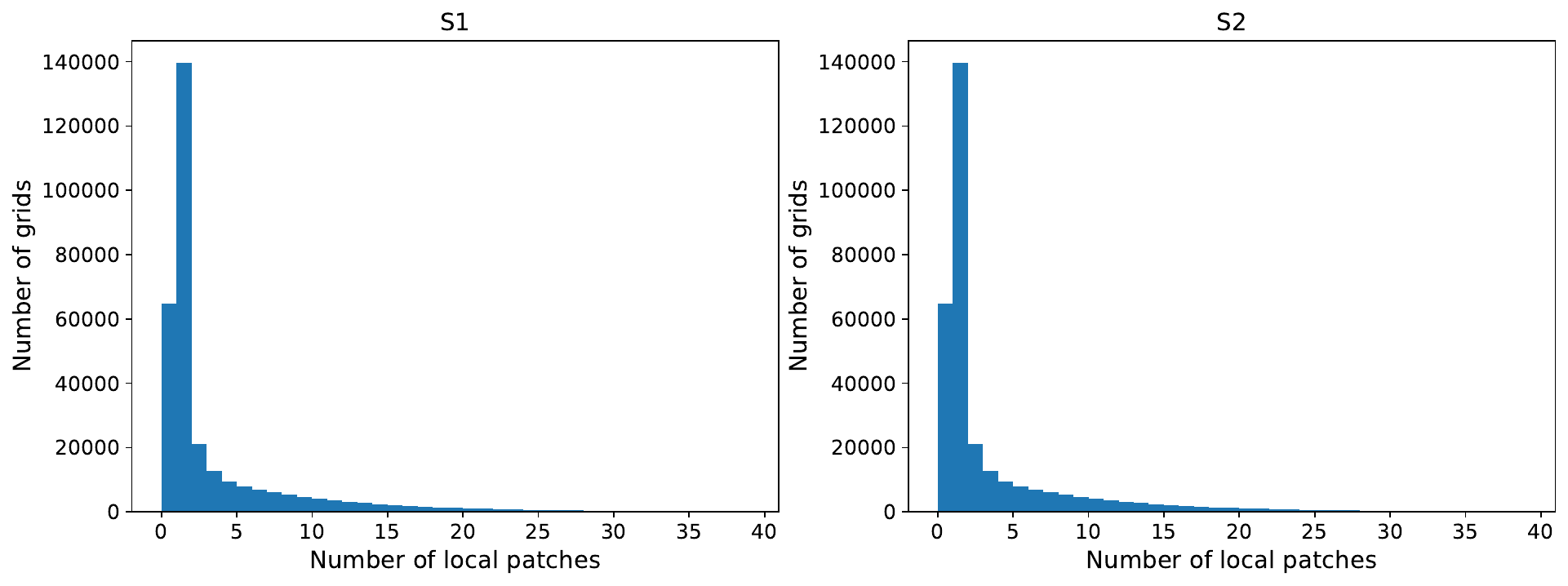}
    \caption{Histogram of local patch numbers for S1 and S2 (union).}
    \label{fig:localpatch_union}
\end{figure*}

\begin{figure*}[ht]
    \centering
    \includegraphics[width=0.8\linewidth]{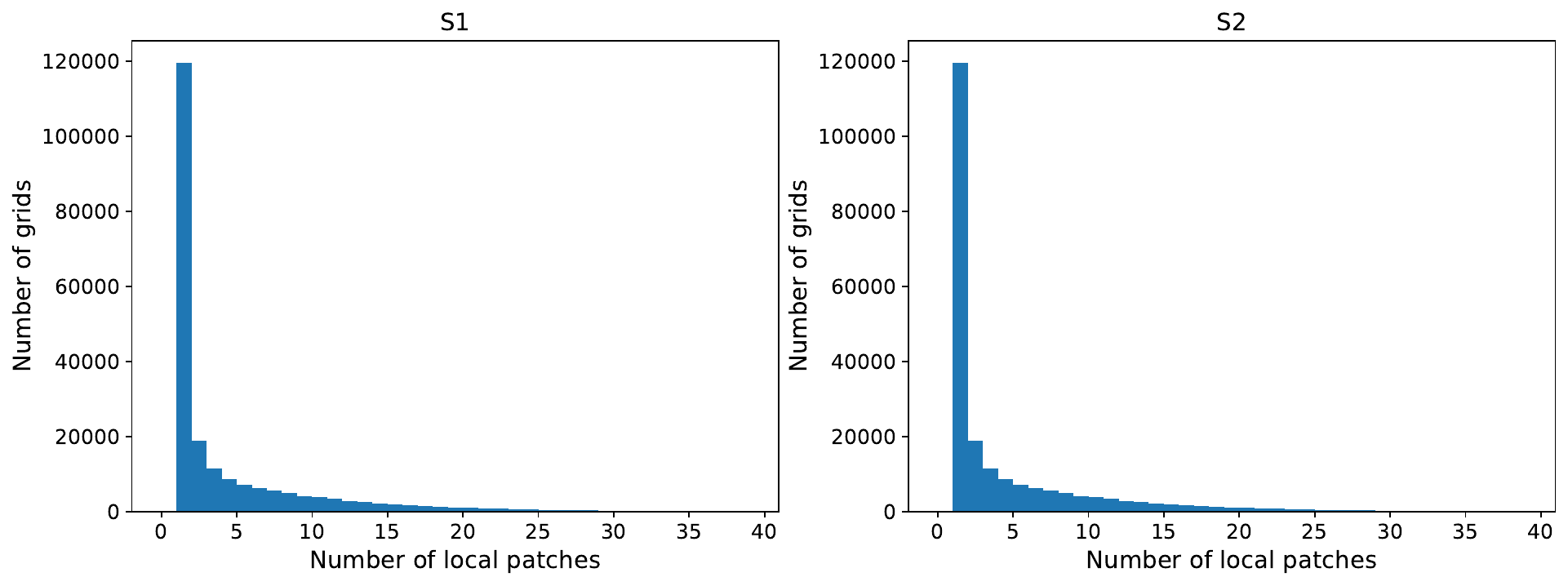}
    \caption{Histogram of local patch numbers for S1 and S2 (joint).}
    \label{fig:localpatch_joint}
\end{figure*}

\paragraph{Statistics of local patches}
\cref{fig:localpatch_union} shows the histograms of the number of local patches across grids for S1/2 in the full datasets (union), and \cref{fig:localpatch_joint} shows the histograms for the joint subset.

\begin{figure*}[ht]
    \centering
    \includegraphics[width=0.8\linewidth]{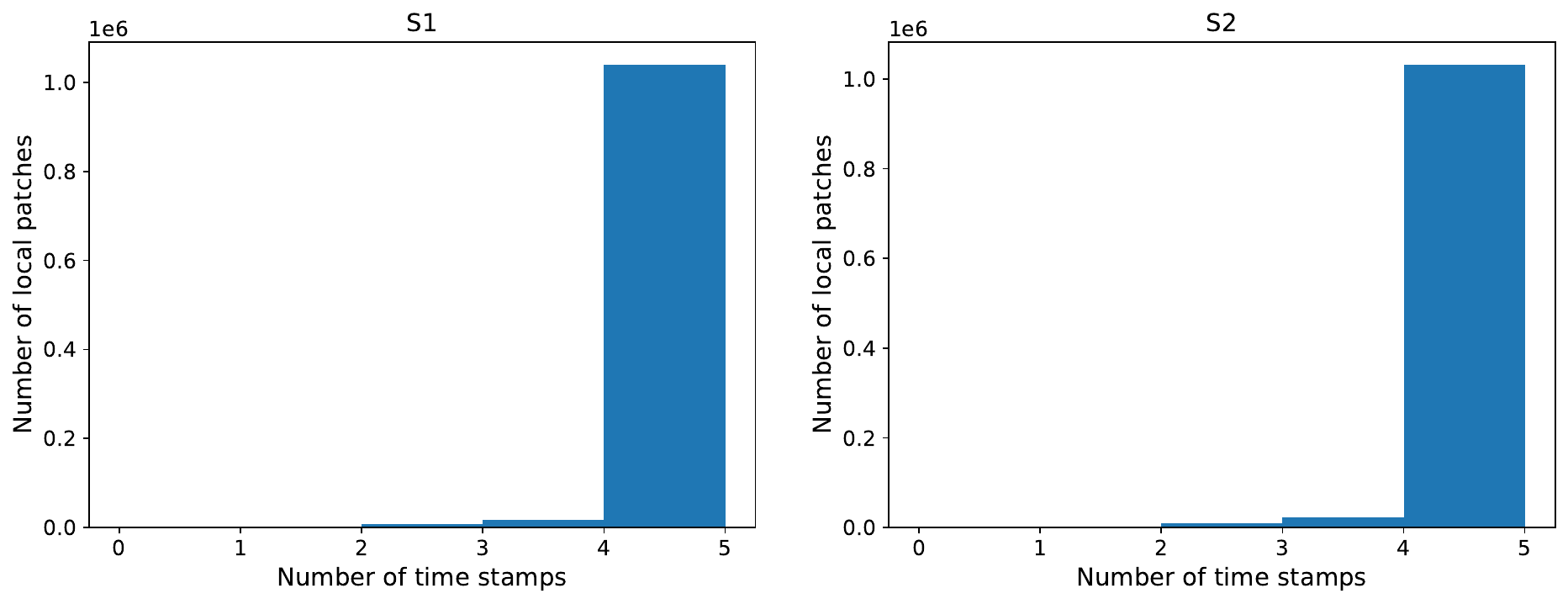}
    
    \caption{Histogram of time series lengths for S1 and S2 (union).}
    \label{fig:time_s12_union}
\end{figure*}

\begin{figure*}[ht]
    \centering
    \includegraphics[width=0.8\linewidth]{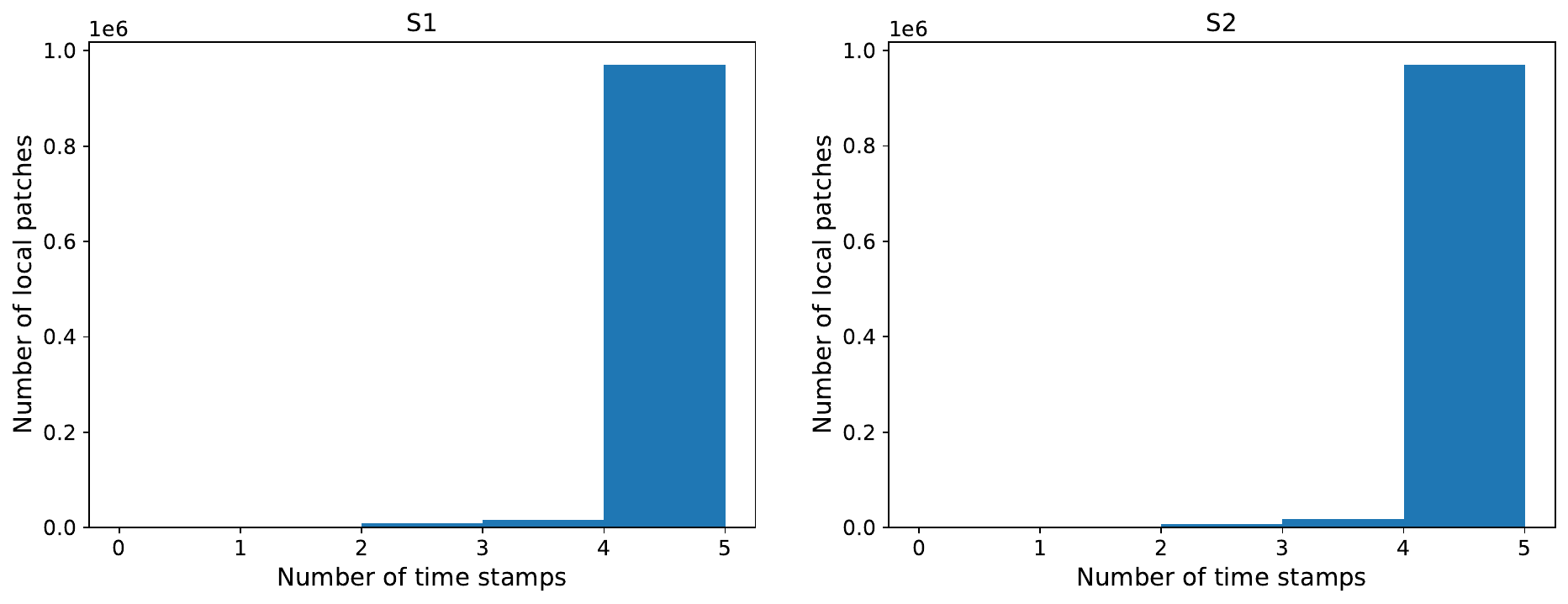}
    \caption{Histogram of time series lengths for S1 and S2 (joint).}
    \label{fig:time_s12_joint}
\end{figure*}

\begin{figure*}[ht]
    \centering
    \includegraphics[width=0.4\linewidth]{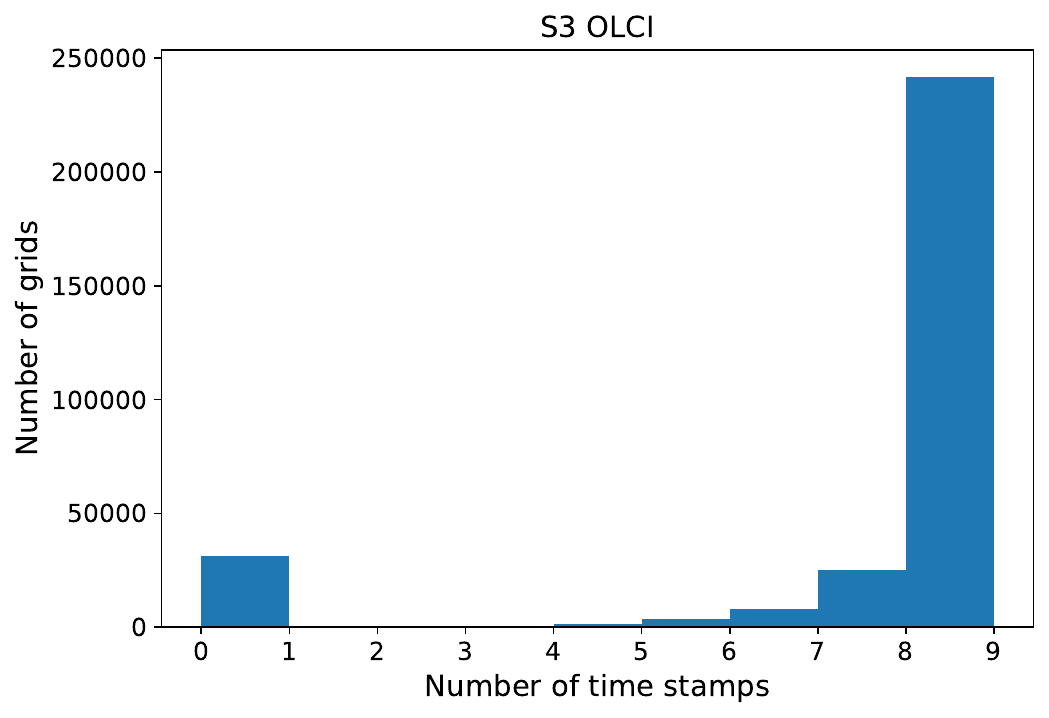}
    \includegraphics[width=0.4\linewidth]{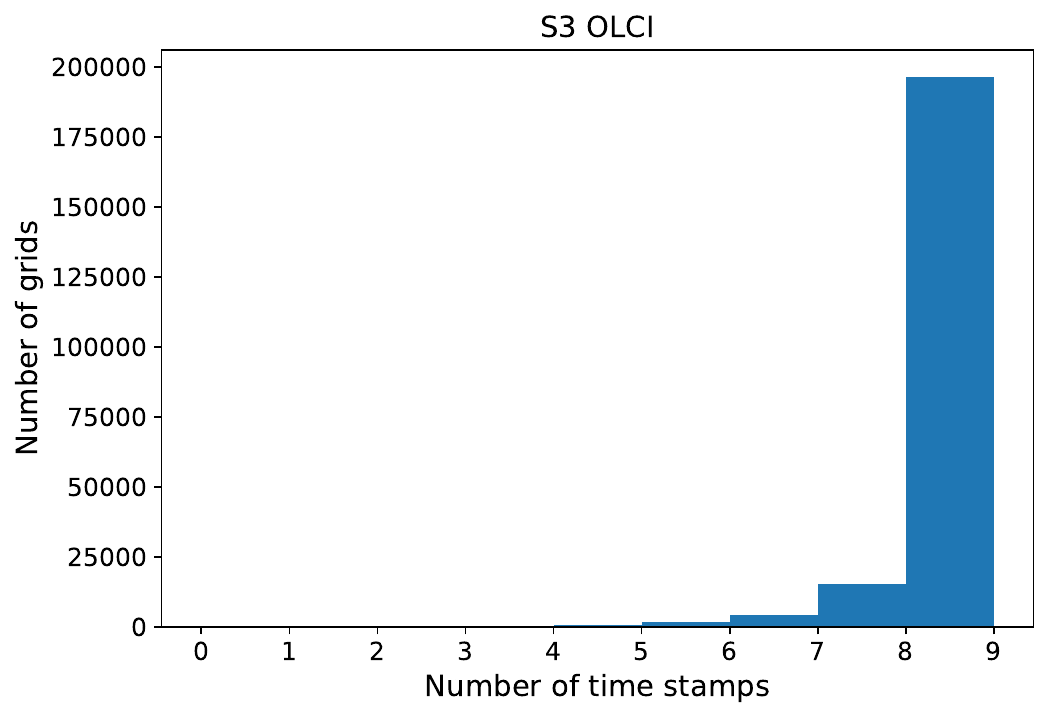}
    \caption{Histogram of time series lengths for S3 (left: union; right: joint).}
    \label{fig:time_s3_union_joint}
\end{figure*}

\begin{figure*}[ht]
    \centering
    \includegraphics[width=0.74\linewidth]{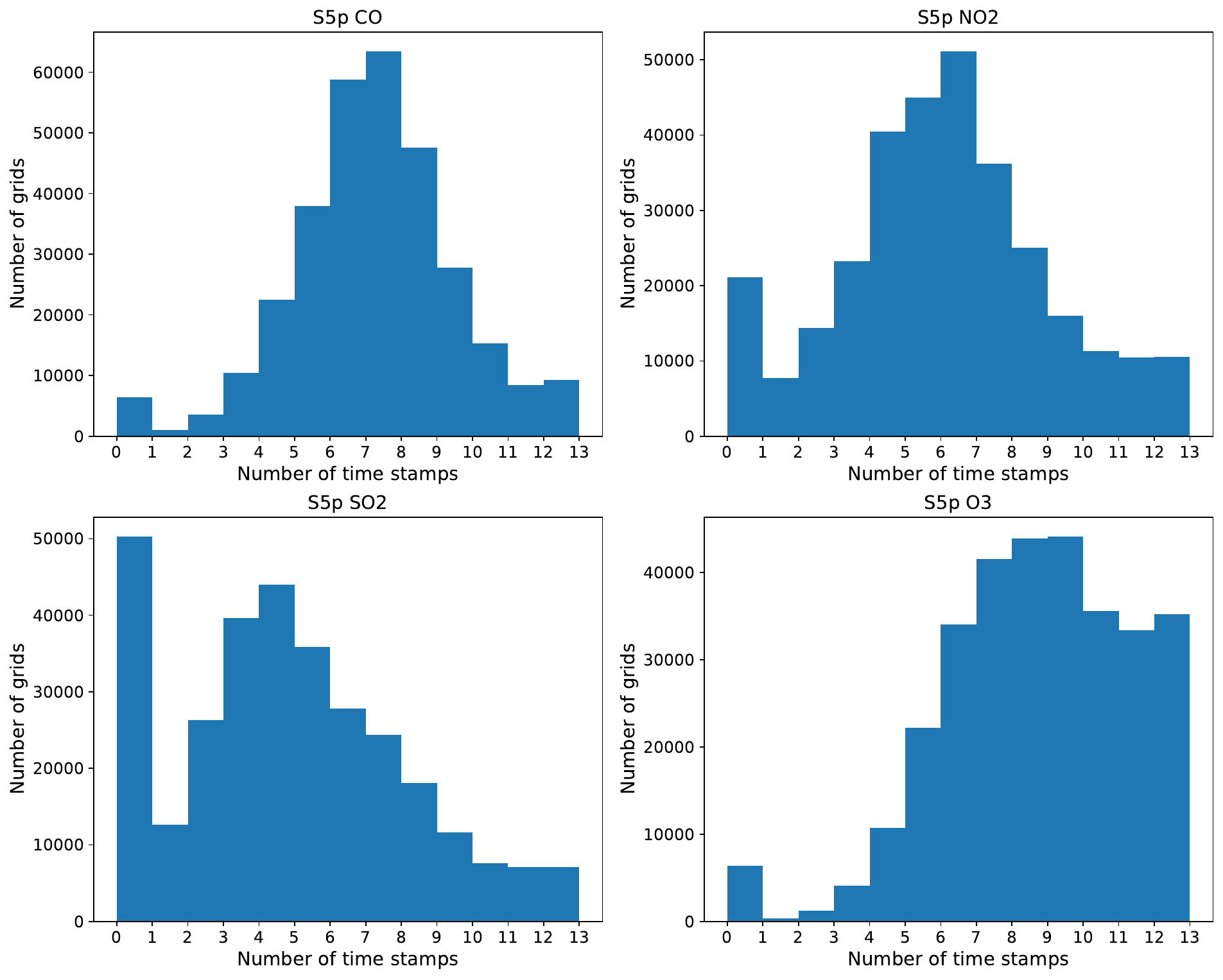}
    \caption{Histogram of time series lengths for S5P (union).}
    \label{fig:time_s5p_union}
\end{figure*}

\begin{figure*}[ht]
    \centering
    \includegraphics[width=0.74\linewidth]{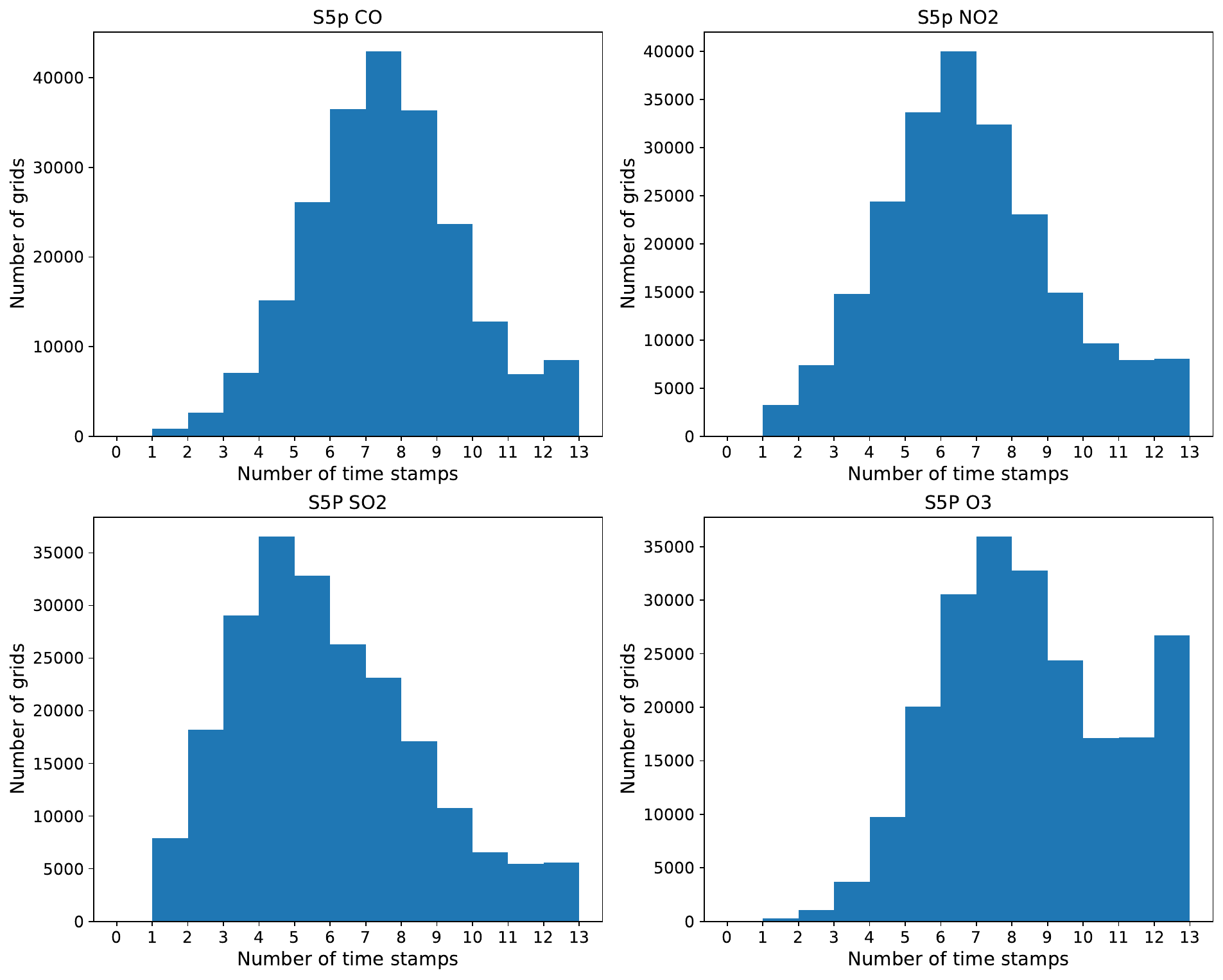}
    \caption{Histogram of time series lengths for S5P (joint).}
    \label{fig:time_s5p_joint}
\end{figure*}

\paragraph{Statistics of time series.}
\cref{fig:time_s12_union} presents the histograms of the time series lengths for S1 and S2 in the full dataset, while \cref{fig:time_s12_joint} shows the corresponding histograms in the joint subset. Similarly, \cref{fig:time_s3_union_joint} (left and right) presents S3 in the full dataset and joint subset, and \cref{fig:time_s5p_union,fig:time_s5p_joint} present S5P in the full dataset and joint subset.

\clearpage
\section{Copernicus-FM}
\label{app:Copernicus-FM}

This section reports more implementation details, analyses, visualizations, and ablation studies for the Copernicus-FM foundation model. Unless explicitly noticed, for most ablation experiments, we pretrain a ViT-Small on a 10K-grid subset of Copernicus-Pretrain for 100 epochs with continual distillation only from DINOv2~\cite{oquab2023dinov2} for efficiency.

\subsection{Subtractive ablation}

The incremental ablation in the main paper demonstrates the design evolution, but does not isolate individual contributions. To complement, we conduct a subtractive ablation in \Cref{tab:ablate1}, showing the benefits of each component regardless of order.

\vspace{-0.5em}
\begin{table}[h]
\centering
\caption{Subtractive ablation. w/o means without.}
\label{tab:ablate1}
\begin{adjustbox}{max width=\linewidth}
\begin{tabular}{lccccc}
\toprule
 & EU-S1 & EU-S2 & EU-RGB & LC-S3 & O3-S5P ($\downarrow$) \\
 \midrule
Copernicus-FM & 81.0 & 89.5 & 78.9 & 90.7 & 811.6 \\ \hdashline
w/o var hypernet & 78.9 & 87.9  & 78.6 & 90.5  & 857.6 \\
w/o metadata & 56.9 & 88.3 & 70.1 & 86.9  & 1556.3 \\
w/o distill & 77.9 & 88.9 & 78.5 & 90.7 & 839.3 \\
\bottomrule
\end{tabular}
\end{adjustbox}
\end{table}

\vspace{-0.5em}
\subsection{Spectral hypernetwork}

We use a unified Fourier encoding~\cite{bodnar2024aurora} to encode wavelengths and bandwidths for all spectral channels, which are added together and serve as input to the spectral hypernetwork to generate patch embedding weights.

\vspace{-0.5em}
\paragraph{Wavelength and bandwidth details}
\cref{tab:wavelengths} lists the detailed wavelength and bandwidth values for each spectral sensor in the Copernicus-Pretrain dataset used during our Copernicus-FM pretraining.

\begin{table}[ht]
    \centering
    \begin{tabular}{lp{7cm}p{7cm}}
        \hline
        Sensor & Wavelengths (nm) & Bandwidths (nm) \\
        \hline
        S1 GRD & 5e7, 5e7 & 1e9, 1e9 \\
        
        S2 TOA & 440, 490, 560, 665, 705, 740, 783, 842, 
             860, 940, 1370, 1610, 2190 & 
             20, 65, 35, 30, 15, 15, 20, 115, 
             20, 20, 30, 90, 180 \\
        
        S3 OLCI & 400, 412.5, 442.5, 490, 510, 560, 620, 665,  
             673.75, 681.25, 708.75, 753.75, 761.25, 764.375,  
             767.5, 778.75, 865, 885, 900, 940, 1020 &  
             15, 10, 10, 10, 10, 10, 10, 10,  
             7.5, 7.5, 10, 7.5, 7.5, 3.75,  
             2.5, 15, 20, 10, 10, 20, 40 \\
        \hline
    \end{tabular}
    \caption{Wavelengths and bandwidths for different spectral sensors in Copernicus-FM pretraining.}
    \vspace{-1em}
    \label{tab:wavelengths}
\end{table}

\begin{figure}[ht]
    \centering
    \includegraphics[width=0.85\linewidth]{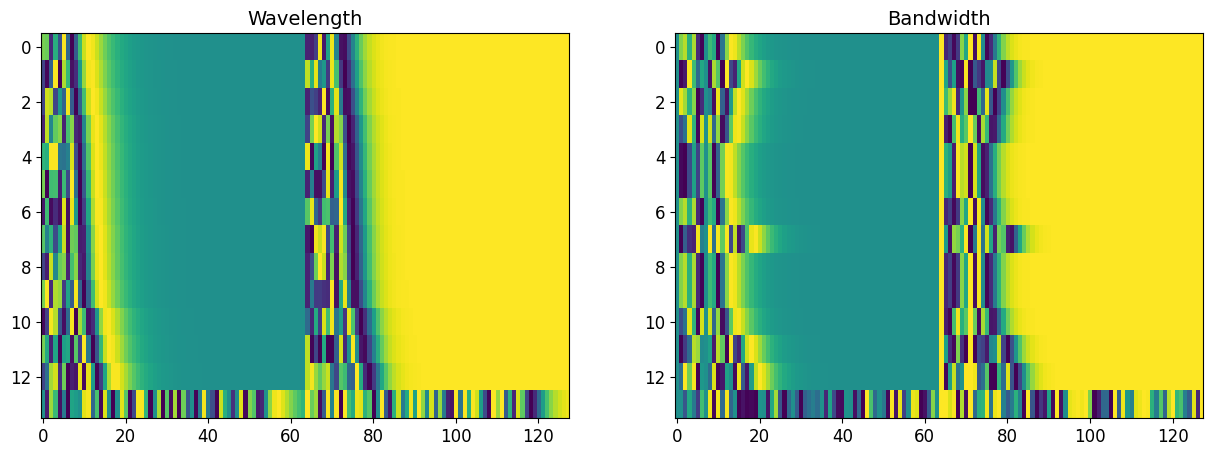}
    \caption{Fourier encoding visualization for wavelengths and bandwidths of S2 and S1.}
    \label{fig:fe-wave}
\end{figure}

\vspace{-0.5em}
\paragraph{Fourier encoding visualization}
\cref{fig:fe-wave} illustrates the Fourier encoded wavelengths and bandwidths (with 128 feature dimensions) for  13 S2 bands and 1 S1 band.

\subsection{Variable hypernetwork}

We use a large language model with general cross-domain knowledge to encode variable names for non-spectral modalities. The resulting variable encodings serve as input to the variable hypernetwork to generate patch embedding weights.

\begin{figure}[ht]
    \centering
    \includegraphics[width=0.7\linewidth]{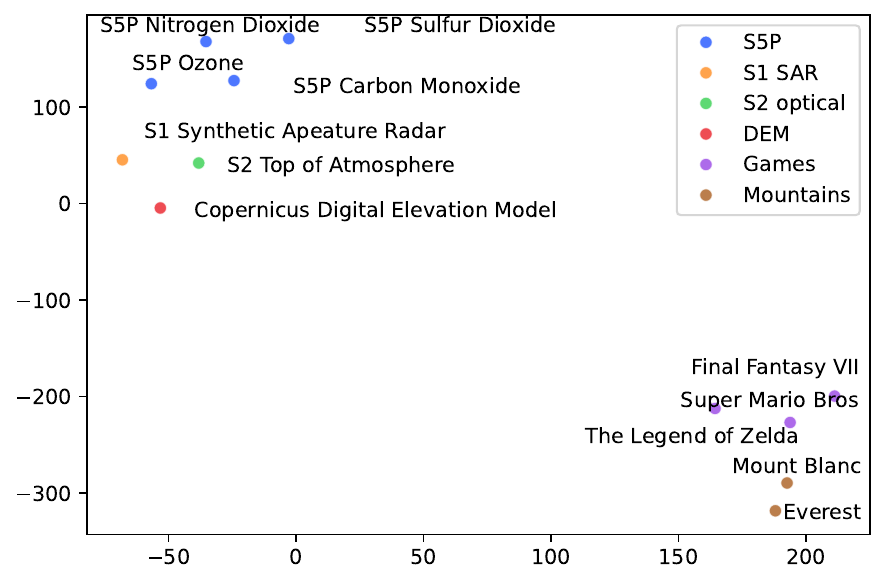}
    \caption{t-SNE visualization of the language encodings of different variable names.}
    \label{fig:tsne-llm}
\end{figure}

\paragraph{Language encoding visualization}
\cref{fig:tsne-llm} presents a t-SNE plot of Llama-3.2-encoded variable names. We compare the variable names in our pretraining dataset with other out-of-domain concepts. The figure indicates that the language model does have meaningful knowledge of these different names --- S5P variables are gathered together, EO modalities are far away from other domains like games or mountains, and concepts within a subdomain are further well clustered.

\paragraph{Ablation on different variable encoding options}

Language encoding maximizes the flexibility to process any variable names, and also maintains semantic relationships between variables. Besides that, random hashing and spectral-sensitivity-guided spectroscopy are two other options to encode different variables. However, compared to language encodings, random hashing loses the flexibility and semantic relationship, while spectroscopy has semantics but lacks flexibility. Quantitatively, we conduct a comparison study in 
\Cref{tab:ablate2}, suggesting the superior and stable performance of language encoding.

\begin{table}[htbp]
\centering
\caption{Variable name encoding ablation.}
\label{tab:ablate2}
\begin{adjustbox}{max width=\linewidth}
\begin{tabular}{lccccc}
\toprule
 & EU-S1 & EU-S2 & EU-RGB & LC-S3 & O3-S5P ($\downarrow$) \\
 \midrule
LLM (LLaMa-3.2-1B) & \textbf{81.0} & \textbf{89.5} & 78.9 & \textbf{90.7} & \textbf{811.6} \\
Random hash & 77.7  & 89.5  & 78.6  & 89.9  & 818.6  \\
Spectroscopy & 80.4 & 89.5 & \textbf{79.5} & 89.9 & 815.7 \\
\bottomrule
\end{tabular}
\end{adjustbox}
\end{table}

% \paragraph{Ablation on different LLMs} Moving further, we wonder how the differences between various LLMs affect the variable encodings' quality. To answer this question, \cref{tab:ablation-llm} conducts additional ablation on different language models to encode the variable names. XXXXXX

% \begin{table}[ht]
% \centering
% \caption{Ablation study on different LLMs for variable name encoding. We report RMSE scores on two air quality regression tasks.}
% \label{tab:ablation-llm}
% \begin{tabular}{lcccl}
% \hline
%  & AQ-NO2-S5P & AQ-O3-S5P \\ \hline
% Llama-3.2-1B \cite{} &  & 811.6 \\
% ClimateBERT \cite{} &  &  \\
% Galactica \cite{} &  &  \\
% SciBERT \cite{} & & \\
% Mistral 7B \cite{} &  &  \\ \hline
% \end{tabular}
% \end{table}

\subsection{Metadata integration}

We use a unified Fourier encoding~\cite{bodnar2024aurora} to integrate metadata as encoding vectors added to the positional encodings.

\begin{figure}[ht]
    \centering
    \includegraphics[width=0.85\linewidth]{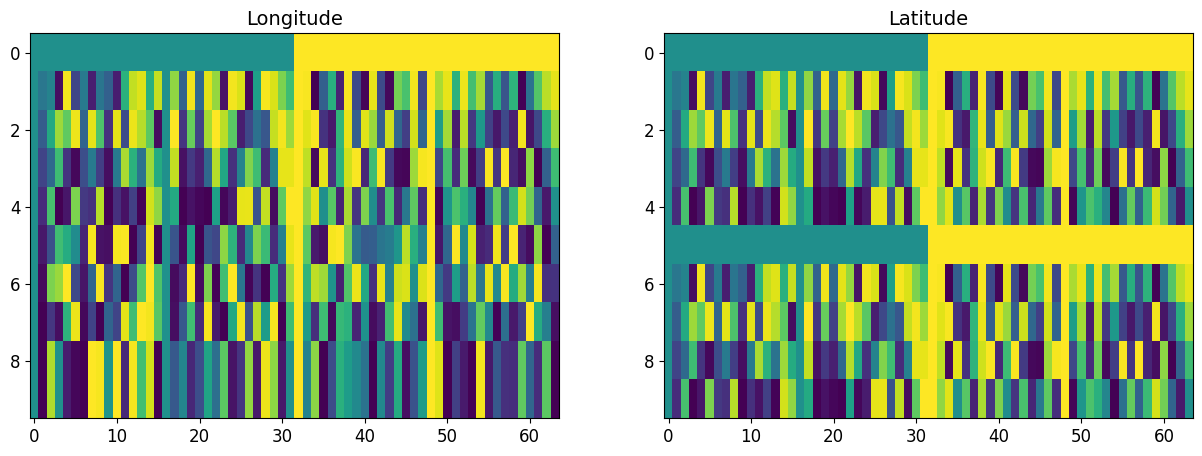}
    \caption{Fourier encoding visualization for geolocation (longitudes and latitudes).}
    \label{fig:fe-loc}
\end{figure}

\begin{figure}[ht]
    \centering
    \includegraphics[width=0.42\linewidth]{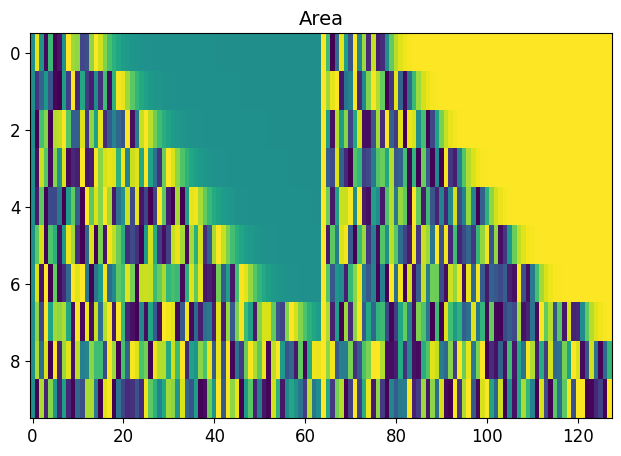}
    \includegraphics[width=0.42\linewidth]{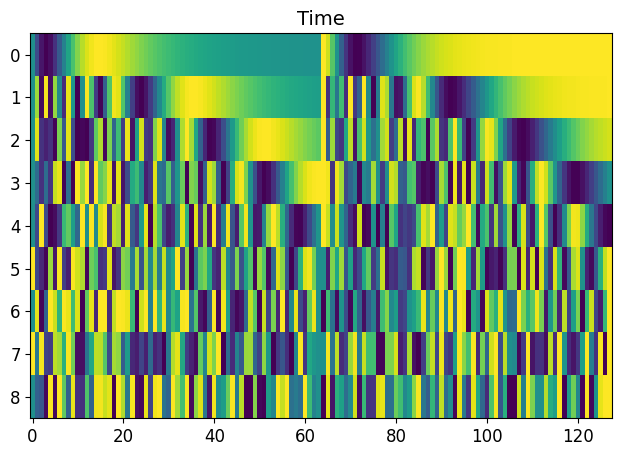}
    \caption{Fourier encoding visualization for area (left) and time (right).}
    \label{fig:fe-area-time}
\end{figure}

\paragraph{Fourier encoding visualization}
\cref{fig:fe-loc,fig:fe-area-time} illustrate the Fourier encoded metadata (location, area, and time) for a few representative example values as below:

\begin{itemize}
    \item location ($\text{lon}+180\degree$): $0\degree, 45\degree, 90\degree, 135\degree, 180\degree, 225\degree, 270\degree, 315\degree, 360\degree, 360\degree$;
    \item location ($\text{lat}+90\degree$): $0\degree, 45\degree, 90\degree, 135\degree, 180\degree, 0\degree, 45\degree, 90\degree, 135\degree, 180\degree$;
    \item area (in km\textsuperscript{2}): 0.1, 1, 10, 100, 1000, 1e4, 1e5, 1e7, 1e8, 5.1e8;
    \item time (in days): 1, 7, 30, 90, 180, 365.25, 730.5, 1826.25, 3652.5.
\end{itemize}

%\vspace{-1em}
\paragraph{Ablation on metadata dropping ratio}
In practice, metadata is not always available as input. We thus randomly drop part of the metadata during pretraining, and use learnable metadata tokens to fill missing metadata encodings. To choose the best metadata dropping probability, \cref{tab:ablation-meta-drop} conducts a corresponding ablation study, where we perform $k$-NN evaluation on three image classification tasks. The table suggests that a relatively high dropping ratio helps improve the model's performance.

\begin{table}[ht]
\centering
\caption{Ablation study on the dropping ratio of metadata. We report overall accuracy with $k$-NN evaluation.}
\label{tab:ablation-meta-drop}
\begin{tabular}{lccc}
\hline
 & EuroSAT-S1 & EuroSAT-S2 & EuroSAT-RGB \\ \hline
metadata (drop 0.1) & 77.8 & 88.7 & 79.9 \\
metadata (drop 0.3) & 73.7 & 86.3 & 77.5 \\
metadata (drop 0.5) & 77.8 & \textbf{89.6} & 78.7 \\
metadata (drop 0.7) & \textbf{81.0} & 89.5 & \textbf{78.9} \\
metadata (drop 0.9) & 78.5 & 88.2 & 74.8 \\ \hline
\end{tabular}
\end{table}

%\vspace{-1em}
\paragraph{Ablation on metadata details}
Moving further, we wonder how much benefit each metadata component brings to the model, as well as how the format of each metadata will affect the performance. To answer these questions, \cref{tab:ablation-meta-type} conducts additional ablation on specific metadata components. Results show that geolocation gives the most significant improvement, followed by area and time. Interestingly, geographic coordinates perform better than Cartesian coordinates despite their distortion in high-latitude regions. Using the area corresponding to the true surface coverage (e.g., cropping and resizing make the true surface coverage smaller) is necessary, without which the performance begins to drop. Using the day of the year and absolute days above one-year-period perform similarly. We use the latter such that it's convenient to extend to long time series in the future.

\begin{table}[ht]
\centering
\caption{Ablation study on the benefits of each metadata type. We report overall accuracy with $k$-NN evaluation. Gray rows are alternative formatting options for the metadata. Performance increases/decreases are compared to the best formatting option of previous metadata.}
\label{tab:ablation-meta-type}
\begin{tabular}{lccc}
\toprule
& EuroSAT-S1 & EuroSAT-S2 & EuroSAT-RGB \\ \midrule
no metadata & 56.9 & 88.3 & 70.1 \\
\rowcolor[HTML]{EFEFEF} + location (x,y,z) & 75.8 \inc{18.9} & 88.7 \inc{0.5} & 73.3 \inc{2.8} \\
/+ location (lon,lat) & 78.2 \inc{21.3} & 88.7 \inc{0.4} & 76.5 \inc{6.5} \\
\rowcolor[HTML]{EFEFEF} + area (raw) & 77.8 \dec{0.4} & 88.1 \dec{0.6} & 73.7 \dec{2.8} \\
/+ area (aug) & 80.3 \inc{2.2} & 89.3 \inc{0.6} & 77.4 \inc{0.8} \\
\rowcolor[HTML]{EFEFEF} + time (dayofyear) & 80.0 \dec{0.3} & 89.5 \inc{0.2} & 78.9 \inc{1.5} \\
/+ time (absolute) & 81.0 \inc{0.7} & 89.5 \inc{0.2} & 78.9 \inc{1.5} \\
\bottomrule
\end{tabular}
\end{table}

\subsection{Pretraining details}

\paragraph{Data}
We pretrain Copernicus-FM on the joint 220K-grid subset of Copernicus-Pretrain, with each grid being one sample unit containing aligned images from all eight modalities. For fast data loading, we convert the raw dataset into webdataset\footnote{\url{https://github.com/webdataset/webdataset}} format, with one grid cell being one minimum sample in the shards. During training, one image from each modality is sampled from one grid cell to construct the input for each iteration. For S1/2, we normalize the image values with channel-wise mean and standard deviation. For S3, we multiply each channel with its corresponding scale factor\footnote{\url{https://developers.google.com/earth-engine/datasets/catalog/COPERNICUS_S3_OLCI}}. For S5P, we use the raw values, and replace NaN pixels with zero. For DEM, we divide the pixel values by 10000. We apply simple data augmentations to each modality, including random resized cropping with scale $[0.2,1.0]$ to its corresponding input size and random horizontal flipping. Each image comes with its metadata, including geolocation (central coordinates in lon/lat), patch area (calculated from GSD and patch size in km\textsuperscript{2}), and time (number of days since a reference date 1970-01-01). The geolocation and patch area are adapted dynamically based on the cropping parameters in data augmentation. Note that despite this adaptation, due to geographical projection the patch area doesn't strictly reflect the surface area, but is accurate enough for our pretraining purpose. While S1/2/3 images have exact acquisition dates, S5P images are monthly mean and DEM doesn't have a specific acquisition date. Therefore, we use the first day of the month for one S5P image, and the first day of the year 2015 for all DEM images.

\paragraph{Model} We use a standard vision Transformer~\cite{dosovitskiy2020image} for the core backbone---e.g., a ViT-Base has 768 hidden dimensions, 12 Transformer blocks, and 12 attention heads. The MLP and attention architectures for the spectral and variable hypernetworks are identical to \citet{xiong2024neural}. For the light decoder to conduct masked image modeling (MIM) pretraining, we also follow \citet{xiong2024neural} and \citet{he2022masked} with 512 hidden dimensions, 8 Transformer blocks, and 16 attention heads. For continual distillation, a projector is used to project the output feature from the student to the frozen teacher model, both after global average pooling.

\paragraph{Loss} We conduct MIM and continual distillation for pretraining. For MIM, we generally follow \citet{he2022masked} to reconstruct masked-out patches for each input modality. The masking ratio is 70\% for all modalities following previous performance studies of MIM in EO~\cite{wang2023ssl4eo,wang2024feature}. For distillation, we distill RGB channels of S2 from frozen DINOv2~\cite{oquab2023dinov2} (ViT-Base with patch size 14) with loss weight 0.1, and full channels of S1 and S2 from frozen SoftCon~\cite{wang2024multi} (ViT-Base with patch size 14) with loss weight 0.2. The former serves as an anchor to control the latent space with general vision knowledge, such that the model can be used on high-resolution or RGB data despite only being pretrained on medium to low resolution Sentinel images. The latter serves as an accelerator to make training converge faster, as well as offering global representation guidance complementary to the main MIM objective. Our preliminary experiments suggest the benefits of the latter S1/2 distillation decrease with longer training times and larger models.

\paragraph{Training} We pretrain Copernicus-FM on 220K Copernicus-Pretrain grids for 100 epochs. The effective batch size is 288. The basic learning rate is 1.5e-4 for batch size 256, and is linearly scaled for varied batch sizes. We warm up the learning rate for 10 epochs, and then apply a cosine decay schedule. We use the AdamW optimizer, with a weight decay of 0.05. One training run takes 512 GPU hours on NVIDIA A100 GPUs, or 128 node hours on one compute node with 4 A100 (40GB).

% \subsection{Additional results on GEO-Bench}

\clearpage
\section{Copernicus-Bench}
\label{app:cobench}

This section presents curation details, more characteristics, and additional visualizations for datasets within Copernicus-Bench, as well as implementation details for the benchmark.

\subsection{Comparison to existing EO benchmarks}
\cref{tab:benchmark-sota} shows a detailed comparison between Copernicus-Bench and several existing EO benchmarks.

\begin{table}[ht]
\centering
\caption{A comparison of existing EO benchmarks.}
\label{tab:benchmark-sota}
\begin{tabular}{llllll}
\toprule
 & \# tasks & task types & modalities & resolution & task range \\ \midrule
SustainBench~\cite{yeh2021sustainbench} & 15 & cls, seg, reg & RGB, MS & 0.6--30~m & surface \\
GEO-Bench~\cite{lacoste2023geo} & 12 & cls, seg & RGB, MS, HS, SAR & 0.1--15~m & surface \\
FoMo-Bench~\cite{bountos2023fomo} & 16 & cls, seg, obj & RGB, MS, HS, SAR & 0.01--60~m & surface \\
PhilEO Bench~\cite{fibaek2024phileo} & 3 & seg, reg & MS & 10~m & surface \\ \midrule
Copernicus-Bench (ours) & 15 & cls, seg, reg, cd & MS, SAR, atmos.\ var. & 10--1000~m & surface, atmosphere \\ \bottomrule
\end{tabular}
\end{table}

\subsection{Benchmark curation}

Copernicus-Bench consists of 15 datasets organized into 3 levels of tasks covering all primary Copernicus Sentinel missions. Among them, nine are derived from existing datasets with permissive licenses, and six are newly curated to fill in the gaps of ML-ready datasets for S3/5P sensors.

\paragraph{Sourced datasets} 
Nine out of 15 datasets in Copernicus-Bench are extracted or adapted from existing datasets:

\begin{itemize}
    \item \textbf{Cloud-S2}: This is a multi-class cloud segmentation dataset derived from CloudSEN12+~\cite{aybar2024cloudsen12+}, one of the largest Sentinel-2 cloud and cloud shadow detection datasets with expert-labeled pixels. We take 25\% samples with high-quality labels, and split them into 1699/567/551 train/val/test subsets.
    \vspace{4pt}
    
    \item \textbf{EuroSAT-S1 and EuroSAT-S2}: These two are multi-class land use/land cover classification datasets taken from EuroSAT~\cite{helber2019eurosat} and EuroSAT-SAR~\cite{wang2024feature}. We follow the train/val/test splits defined in \citet{neumann2019domain} with 16200/5400/5400 train/val/test images. Images of the two datasets are one-to-one paired, thus they can also be combined to serve as a multimodal image classification dataset. These two datasets do not have time metadata.
    \vspace{4pt}
    
    \item \textbf{BigEarthNet-S1 and BigEarthNet-S2}: These two datasets are sourced from BigEarthNet-v2~\cite{clasen2024reben}, a large-scale S1/2 dataset for multilabel land use/land cover classification. We sample a 5\% subset (11894/6117/5991 images) from each of the official train/val/test splits, respectively. Images from the two datasets are again one-to-one paired, thus they can be combined to serve as a multimodal multilabel image classification dataset. In addition, each S1/2 image pair has a corresponding land cover map in 100~m resolution, thus they can also be used as pixel-level segmentation datasets.
    \vspace{4pt}

    \item \textbf{DFC2020-S1 and DFC2020-S2}: These two are land use/land cover segmentation datasets derived from the IEEE GRSS Data Fusion Contest 2020 (DFC2020)~\cite{rha7-m332-19}. We take S1/2 images and 10~m-resolution labels from the original test set, and further split them into 3156/986/986 train/val/test subsets. Again, images from S1 and S2 datasets are one-to-one paired, thus they can be combined to serve as a multimodal semantic segmentation dataset. These two datasets do not have geolocation and time metadata.
    \vspace{4pt}

    \item \textbf{Flood-S1}: This is a flood segmentation dataset extracted from a large flood mapping dataset Kuro Siwo~\cite{bountos2023kuro}. The original dataset is organized according to various flooding events around the globe. We take a random subset of samples that contain at least the water class to construct 3000/1000/1000 train/val/test subsets. Each sample contains two pre- and one post-event S1 SAR image, forming a time-series segmentation or a change detection dataset. By default, we use one pre-event and one post-event image in Copernicus-Bench.
    \vspace{4pt}

    \item \textbf{LCZ-S2}: This is a multi-class scene classification dataset derived from So2Sat-LCZ42~\cite{zhu2020so2sat}, a large-scale local climate zone classification dataset. We randomly select 25K samples from the training set of the "cultural-10" version to construct new 15000/5000/5000 train/val/test subsets. The original data contains also S1 data, thus this dataset can also be extended to an S1 task and a multimodal task. This dataset does not have geolocation and time metadata.
    \vspace{4pt}
    
\end{itemize}

\paragraph{New datasets}
Six of 15 datasets in Copernicus-Bench are newly curated:

\begin{itemize}
    \item \textbf{Cloud-S3}: This is a cloud segmentation dataset with raw images from Sentinel-3 OLCI and labels from the IdePix~\cite{wevers_2022_6517333} classification algorithm. We first download a few large cloudy S3 tiles (about $4800\times400$ pixels) distributed across the globe, and then apply the IdePix algorithm using the ESA SNAP toolbox to get multi-class cloud masks. After that, we manually check the quality of the generated masks, filter out low-quality tiles, and get seven big tiles with high-quality cloud labels. Next, we remap the label IDs, georeference the tiles to GeoTIFFs, and use GDAL to crop the large tiles into small patches with size $256\times256$ pixels. We remove boundary patches filled with NaN pixels, and split all high-quality patches into 1197/399/399 train/val/test subsets. The class names for the cloud masks are: invalid, clear, cloud-sure, cloud-ambiguous, cloud-shadow, and snow-ice, of which ``invalid'' should be ignored during training. Apart from the multi-class labels, for each image we also have one binary cloud mask. Therefore, the Cloud-S3 dataset can serve as both a multi-class and a binary segmentation dataset.
    \vspace{4pt}

    \item \textbf{LC100Cls-S3 and LC100Seg-S3}: These two datasets are based on Sentinel-3 OLCI images and CGLS-LC100~\cite{Buchhorn2020Copernicus} land cover maps. CGLS-LC100 is a product in the Copernicus Global Land Service (CGLS) portfolio and delivers a global 23-class land cover map at 100~m spatial resolution. We pick the map product for 2019, and sample and download S3 images and LC100 labels for about 10K locations across the globe using GEE. For each location, we download a land cover map with about $288\times288$ pixels, and four seasonal S3 OLCI images each with about $96\times96$ pixels (300~m resolution). Despite using bright pixel percentage to simulate cloud filtering, the resulting images still contain a large volume of clouds. To tackle this issue, we train a cloud detection model based on the previously introduced Cloud-S3 dataset and filter out model-detected cloudy images. After a final quality check, we get about 8K samples each with a land cover map and a time series of S3 images. We divide them into 5181/1727/1727 train/val/test subsets to construct the LC100Seg-S3 dataset. For LC100Cls-S3, we integrate multi-label annotations from the land cover maps for each sample, constructing a multilabel classification dataset. Note that the number of S3 time stamps for different samples may differ because of the cloud filtering process. Apart from the time series, we also pre-select one image for each sample, constructing the ``static'' version of LC100Cls-S3 and LC100Seg-S3, which is the default mode in Copernicus-Bench.
    \vspace{4pt}

    \item \textbf{Biomass-S3}: This regression dataset is based on Sentinel-3 OLCI images and CCI biomass~\cite{Santoro2024}. The biomass product is part of the European Space Agency's Climate Change Initiative (CCI) program and delivers global forest above-ground biomass at 100~m spatial resolution. We pick the product for 2020, and the layer of above ground biomass (AGB, unit: tons/ha, i.e. Mg/ha) as regression ground truth, which is defined as the mass, expressed as oven-dry weight of the woody parts (stem, bark, branches and twigs) of all living trees excluding stump and roots. We sample representative regions across the globe, and download corresponding S3 images (one for each season) from GEE and biomass maps from the CCI open data portal. We crop the S3 images into patches with about $96\times96$ pixels (300~m resolution), and the corresponding biomass maps into patches with about $288\times288$ pixels. Similar to LC100Cls-S3 and LC100Seg-S3, the resulting S3 images contain a large volume of clouds, thus we use again the cloud detection model to filter out cloudy images. After a final quality check, we acquire 5K samples each with a biomass map and a time series of S3 images. We divide them into 3000/1000/1000 train/val/test subsets to construct the Biomass-S3 dataset. Note that the number of S3 time stamps for different samples may differ because of the cloud filtering process. Apart from the time series, we also pre-select one image for each sample, constructing the ``static'' version of Biomass-S3, which is also the default mode in Copernicus-Bench.
    \vspace{4pt}

    \item \textbf{AQ-NO2-S5P and AQ-O3-S5P}: These two regression datasets are based on Sentinel-5P NO\textsubscript{2} and O\textsubscript{3} images and EEA air quality data products~\cite{Horálek2024}. The European Environment Agency (EEA) air quality product provides values for the human health related indicators of air pollutants at 1~km\textsuperscript{2} grid covering the whole Europe, combining monitoring air quality data in a ``regression-interpolation-merging-mapping'' methodology and the observational values of the air quality monitoring stations used in the interpolation. We pick the products in 2021 for NO\textsubscript{2} (annual average concentration) and Ozone (O\textsubscript{3}, 93.2 percentile of maximum daily 8-hour means, SOMO35) as regression ground truth. We sample and download S5P NO\textsubscript{2} (``tropospheric\_NO2\_column\_number\_density'') and O\textsubscript{3} (``O3\_column\_number\_density'') images from GEE, and EEA NO\textsubscript{2} and O\textsubscript{3} maps from EEA datahub\footnote{\url{https://www.eea.europa.eu/en/datahub}}. We use a sample patch size of about $56\times56$ pixels for both S5P and EEA. For S5P, we download two versions: 1) annual mean, and 2) seasonal mean for each season. After filtering out NaN patches and a final quality check, we get 1480/493/494 train/val/test samples for both NO\textsubscript{2} and O\textsubscript{3}, each with an ``annual'' mode of 1 S5P image and a ``seasonal'' mode of 4 S5P images. ``Annual'' is the default mode in Copernicus-Bench.
    %\vspace{5pt}
    
\end{itemize}

\subsection{Benchmark characteristics}

\begin{figure}[ht]
    \centering
    \includegraphics[width=0.45\linewidth]{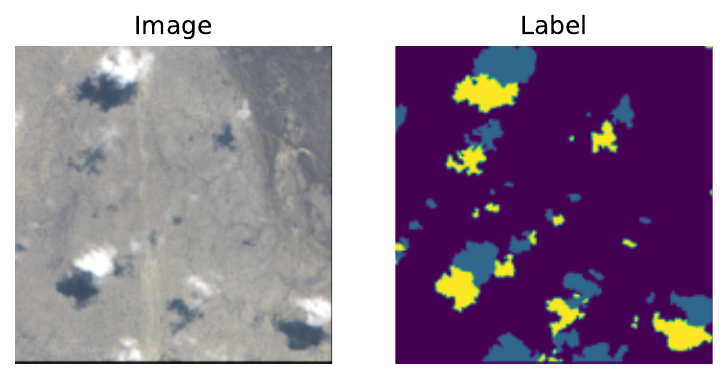}
    \includegraphics[width=0.45\linewidth]{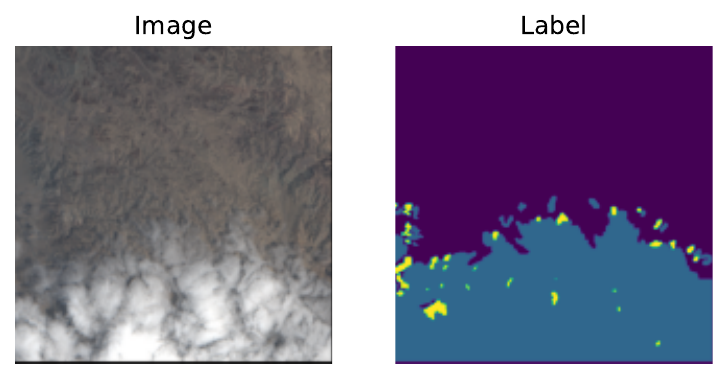}
    \caption{Copernicus-Bench-Cloud-S2.}
    \label{fig:vis-clouds2}
\end{figure}

\begin{figure}[ht]
    \centering
    \includegraphics[width=0.45\linewidth]{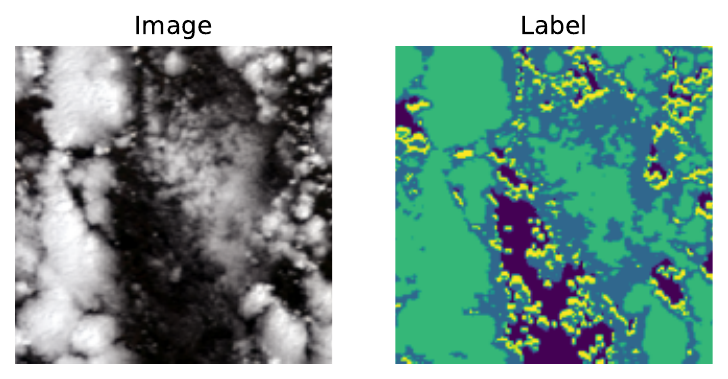}
    \includegraphics[width=0.45\linewidth]{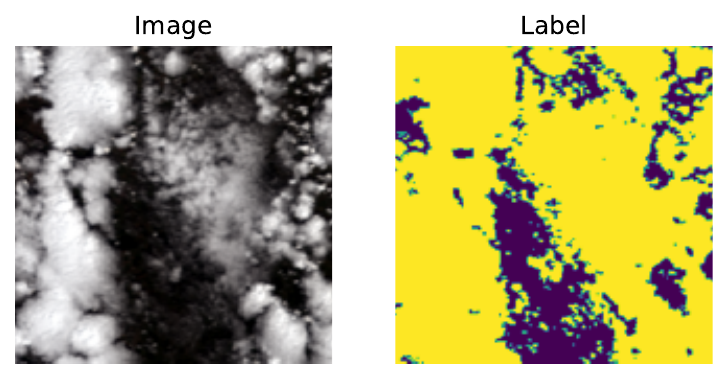}
    \caption{Copernicus-Bench-Cloud-S3. Left: ``multi-class'' mode. Right: ``binary'' mode.}
    \label{fig:vis-clouds3}
\end{figure}

\begin{figure}[ht]
    \centering
    \includegraphics[width=0.45\linewidth]{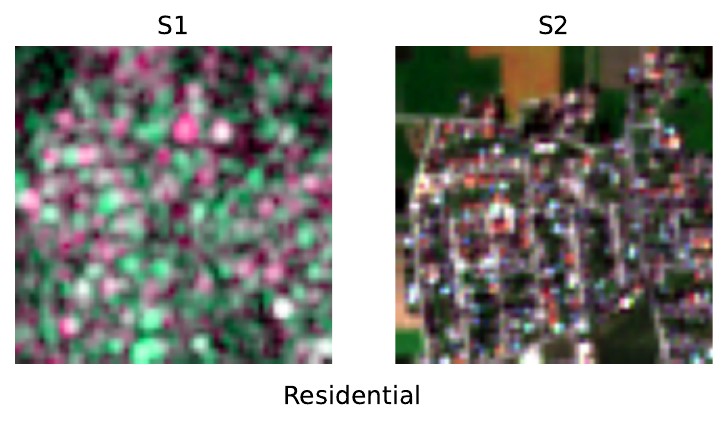}
    \includegraphics[width=0.45\linewidth]{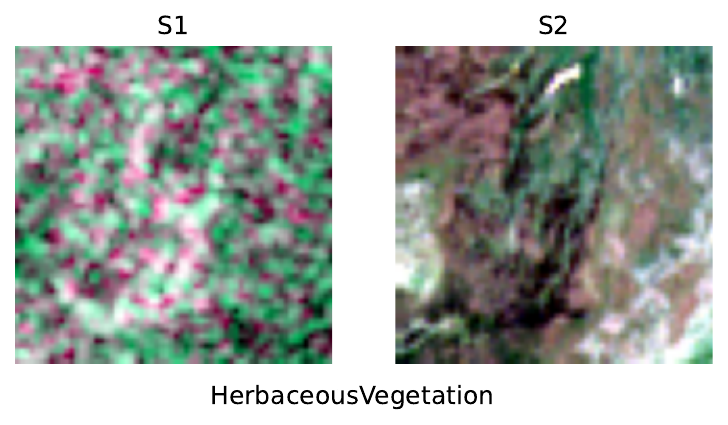}
    \caption{Copernicus-Bench-EuroSAT-S1 and Copernicus-Bench-EuroSAT-S2.}
    \label{fig:vis-eu}
\end{figure}

\begin{figure}[ht]
    \centering
    \includegraphics[width=0.45\linewidth]{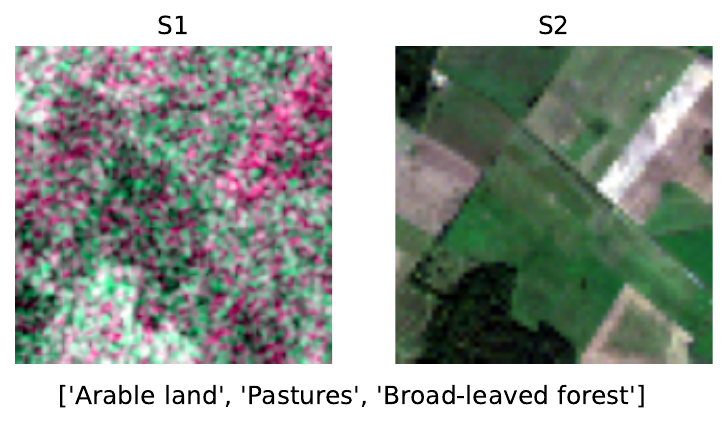}
    \includegraphics[width=0.45\linewidth]{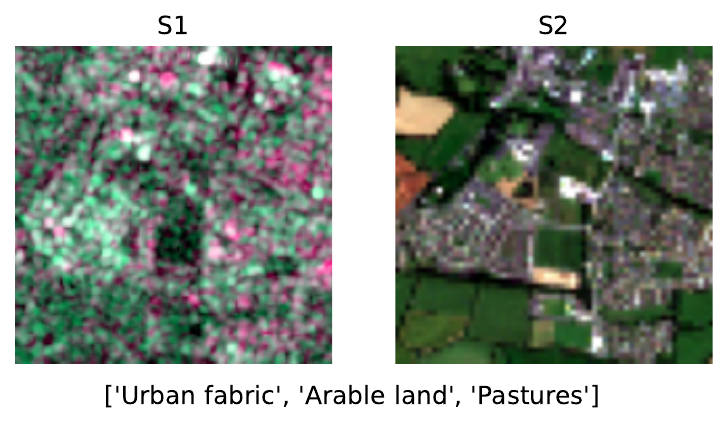}
    \caption{Copernicus-Bench-BigEarth-S1 and Copernicus-Bench-BigEarth-S2.}
    \label{fig:vis-ben}
\end{figure}

\begin{figure}[ht]
    \centering
    \includegraphics[width=0.8\linewidth]{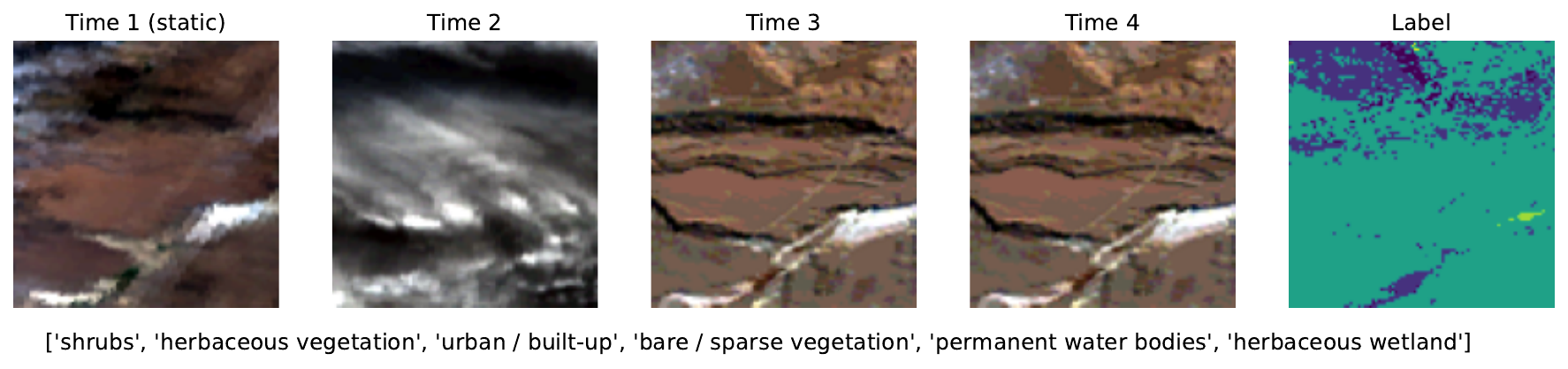}
    \caption{Copernicus-Bench-LC100Cls-S3 and Copernicus-Bench-LC100Seg-S3. By default we pick one image per time series as ``static'' mode.}
    \label{fig:vis-lc100}
\end{figure}

\begin{figure}[ht]
    \centering
    \includegraphics[width=0.45\linewidth]{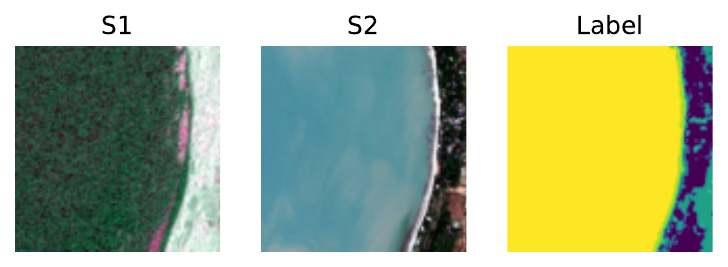}
    \includegraphics[width=0.45\linewidth]{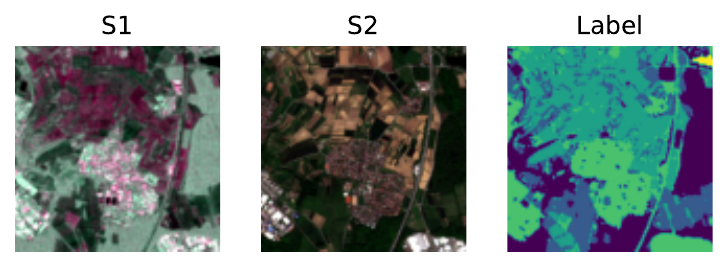}
    \caption{Copernicus-Bench-DFC2020-S1 and Copernicus-Bench-DFC2020-S2.}
    \label{fig:vis-dfc}
\end{figure}

\begin{figure}[ht]
    \centering
    \includegraphics[width=0.8\linewidth]{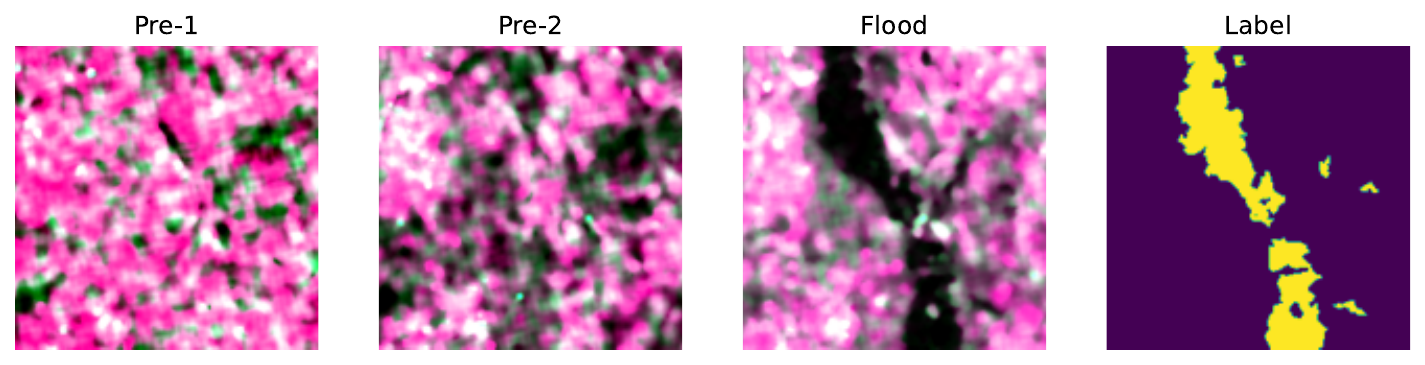}
    \caption{Copernicus-Bench-Flood-S1.}
    \label{fig:vis-flood}
\end{figure}

\begin{figure}[ht]
    \centering
    \includegraphics[width=0.15\linewidth]{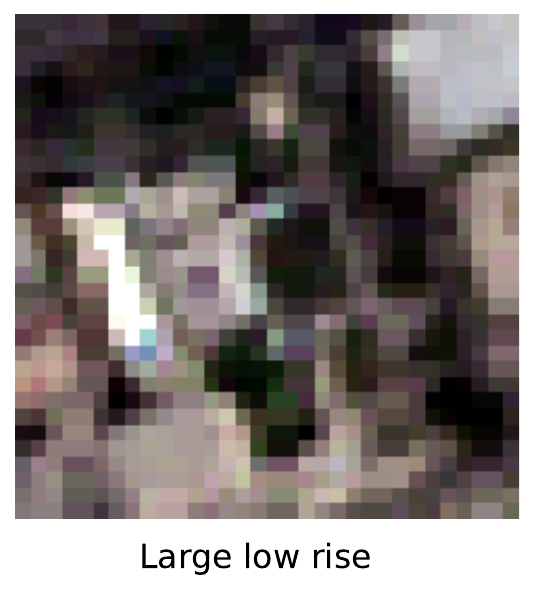}
    \includegraphics[width=0.15\linewidth]{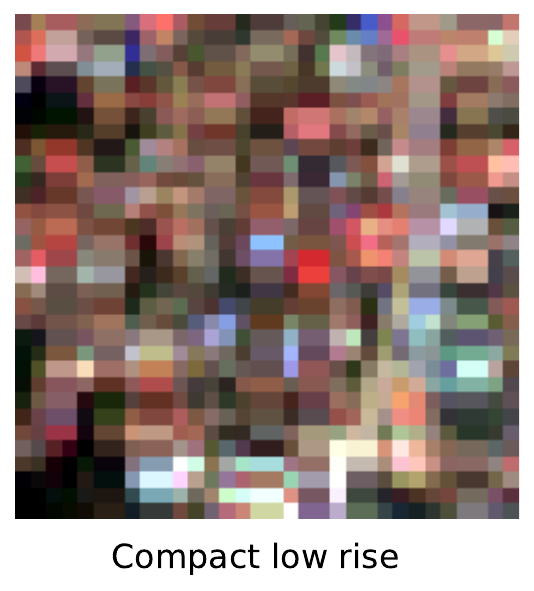}
    \includegraphics[width=0.15\linewidth]{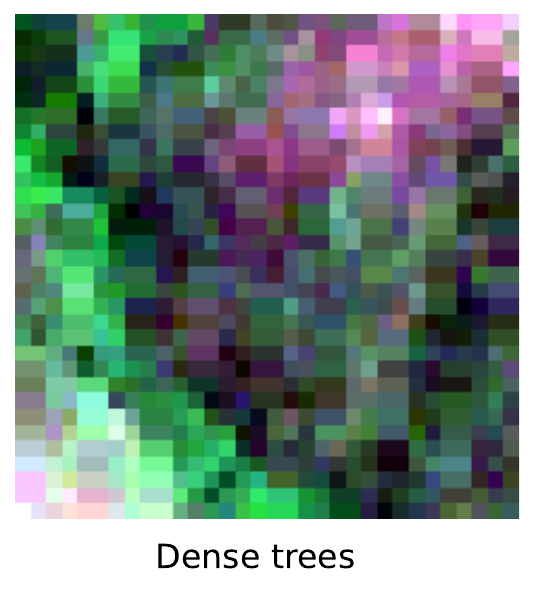}
    \includegraphics[width=0.15\linewidth]{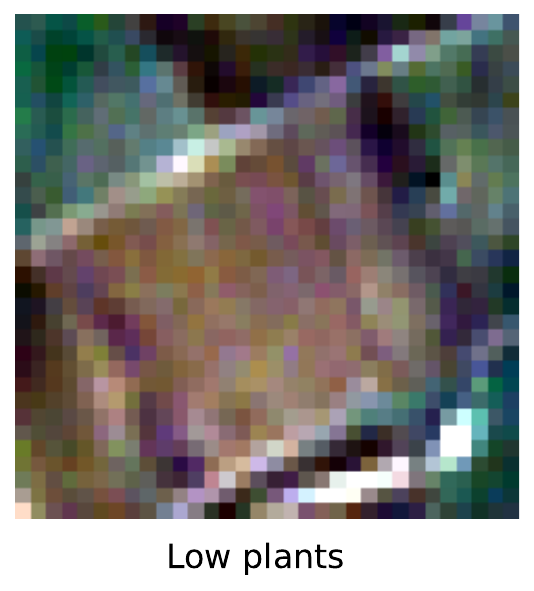}
    \includegraphics[width=0.15\linewidth]{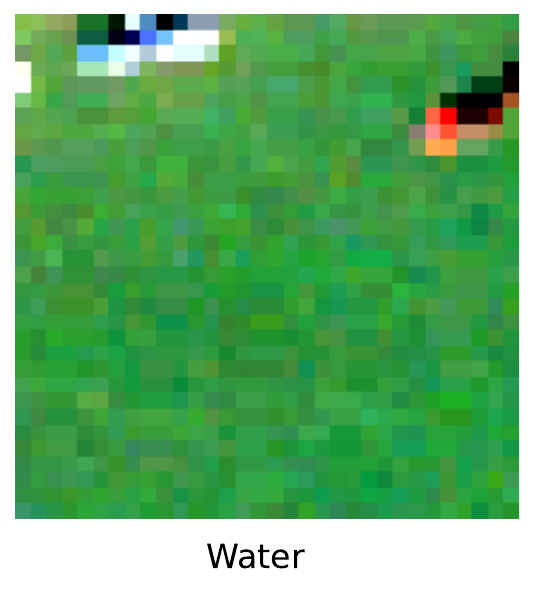}
    \caption{Copernicus-Bench-LCZ-S2.}
    \label{fig:vis-lcz}
\end{figure}

\begin{figure}[ht]
    \centering
    \includegraphics[width=0.9\linewidth]{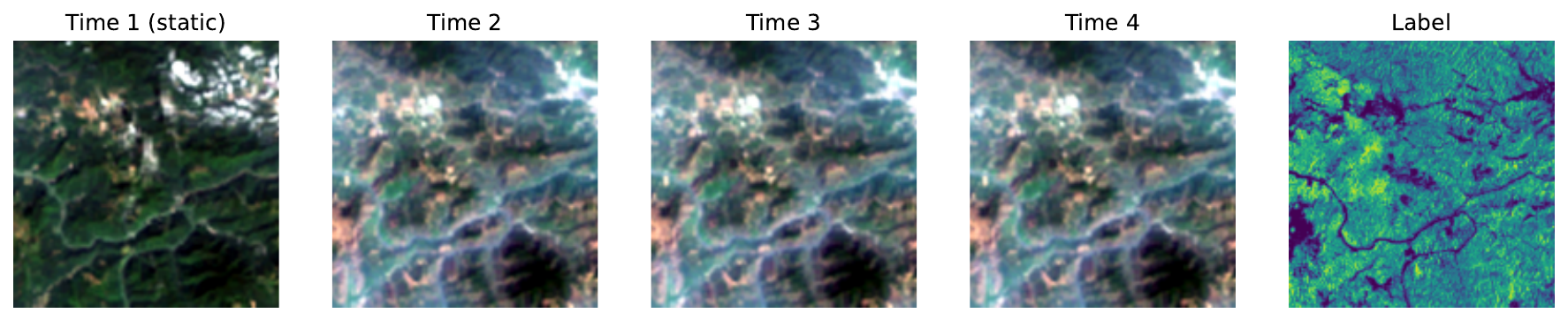}
    \caption{Copernicus-Bench-Biomass-S3. By default we pick one image per time series as ``static'' mode.}
    \label{fig:vis-biomass}
\end{figure}

\begin{figure}[ht]
    \centering
    \includegraphics[width=0.75\linewidth]{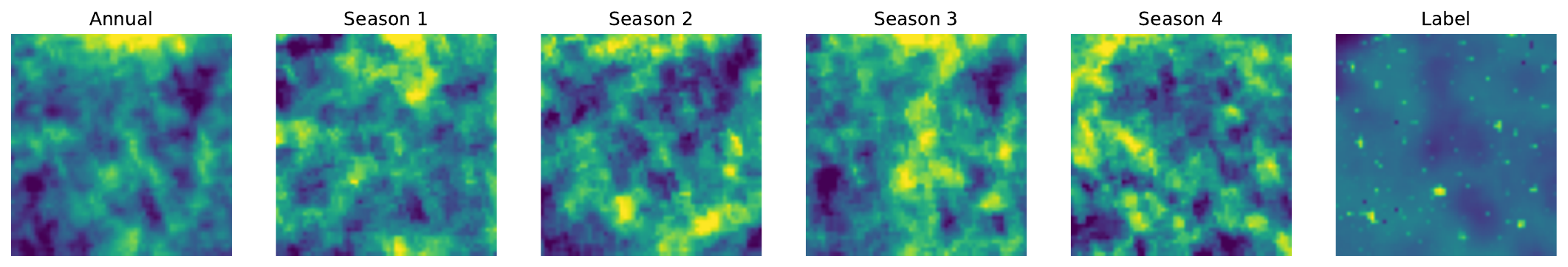}
    \includegraphics[width=0.75\linewidth]{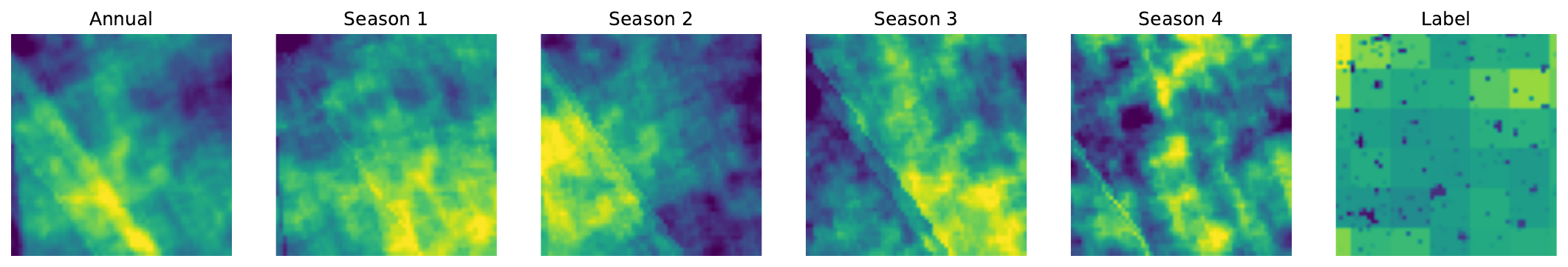}    
    \caption{Copernicus-Bench-AQ-NO2-S5P and Copernicus-Bench-AQ-O3-S5P. By default we pick the ``annual'' mode.}
    \label{fig:vis-aq}
\end{figure}

\paragraph{Example visualization}
\cref{fig:vis-clouds2,fig:vis-clouds3,fig:vis-eu,fig:vis-ben,fig:vis-lc100,fig:vis-dfc,fig:vis-flood,fig:vis-lcz,fig:vis-biomass,fig:vis-aq} visualize some examples for each dataset in Copernicus-Bench.

% \clearpage

%\vspace{-1em}

%\clearpage

\begin{figure}[ht]
    \centering
    \includegraphics[width=0.455\linewidth]{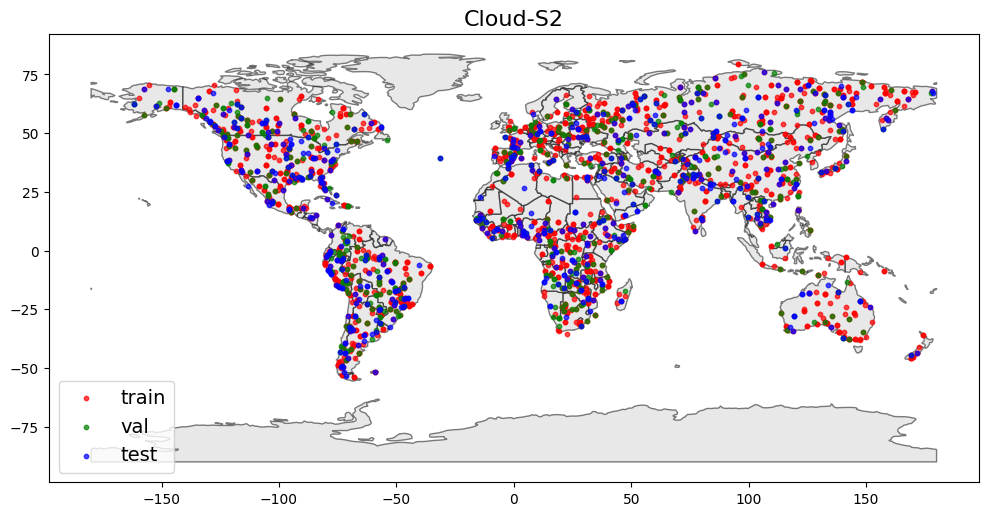}
    \includegraphics[width=0.455\linewidth]{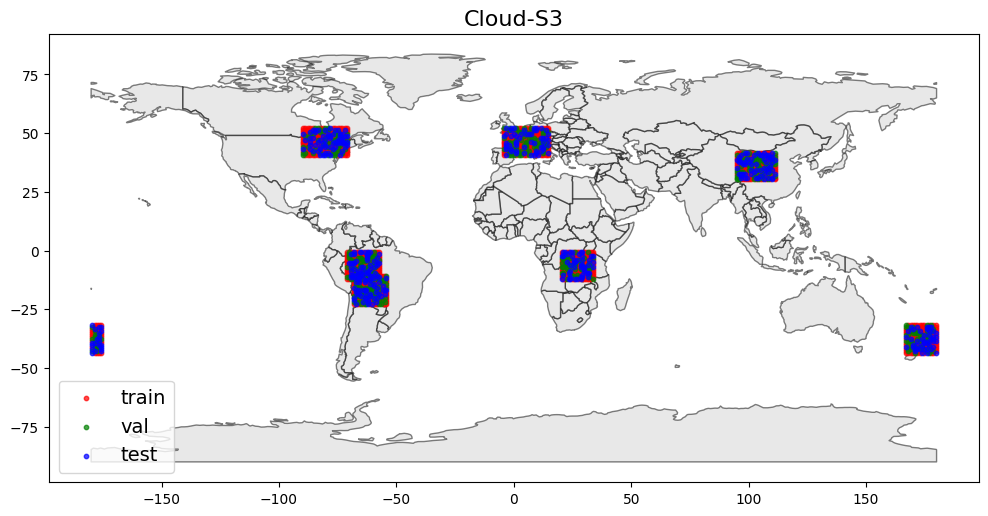}
    \includegraphics[width=0.455\linewidth]{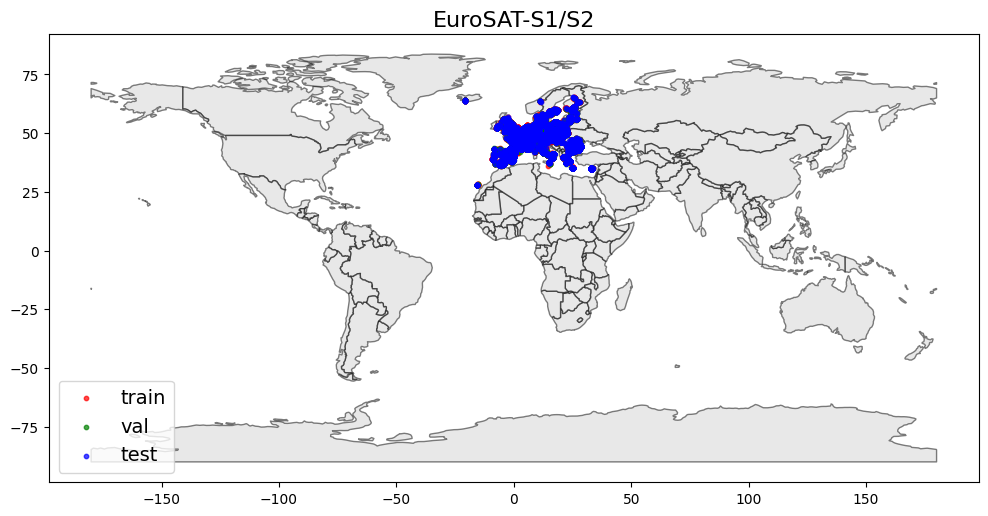}
    \includegraphics[width=0.455\linewidth]{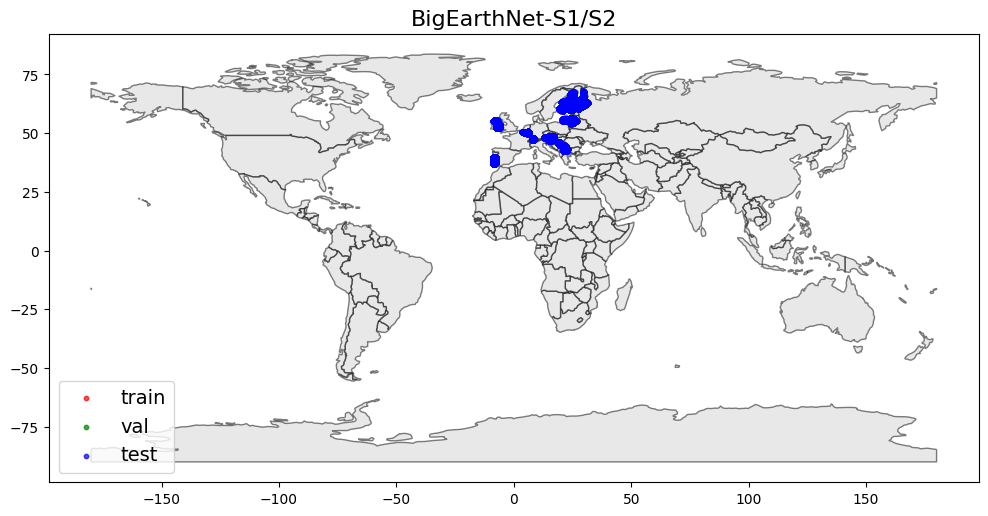}
    \includegraphics[width=0.455\linewidth]{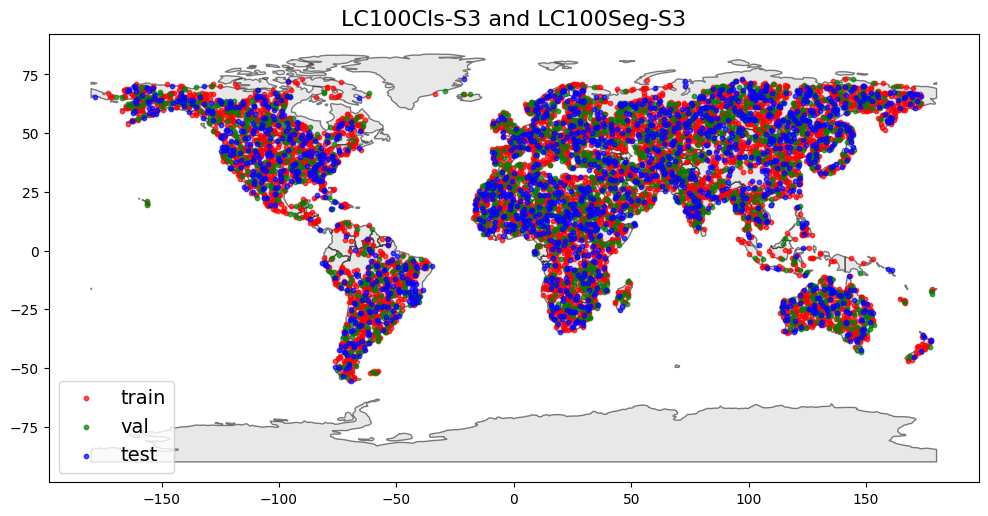}
    \includegraphics[width=0.455\linewidth]{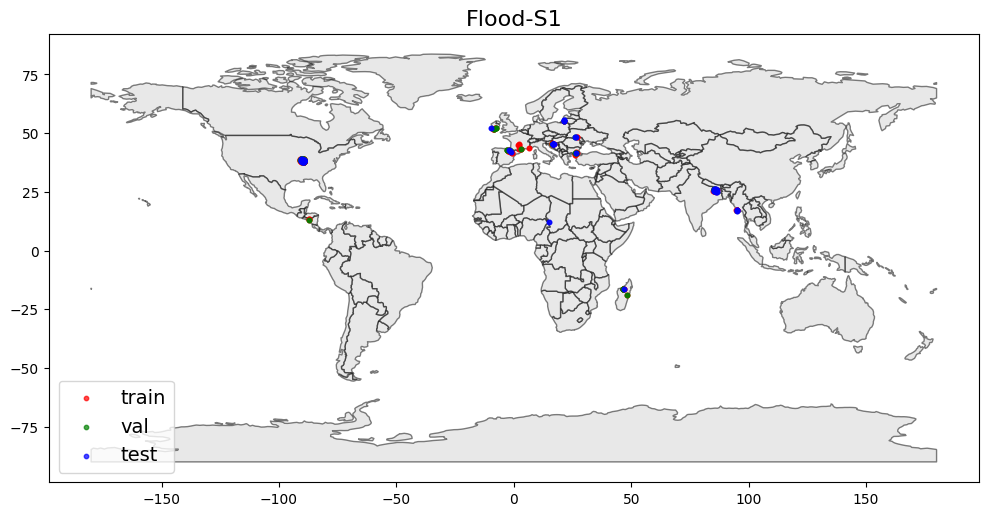}
    \includegraphics[width=0.455\linewidth]{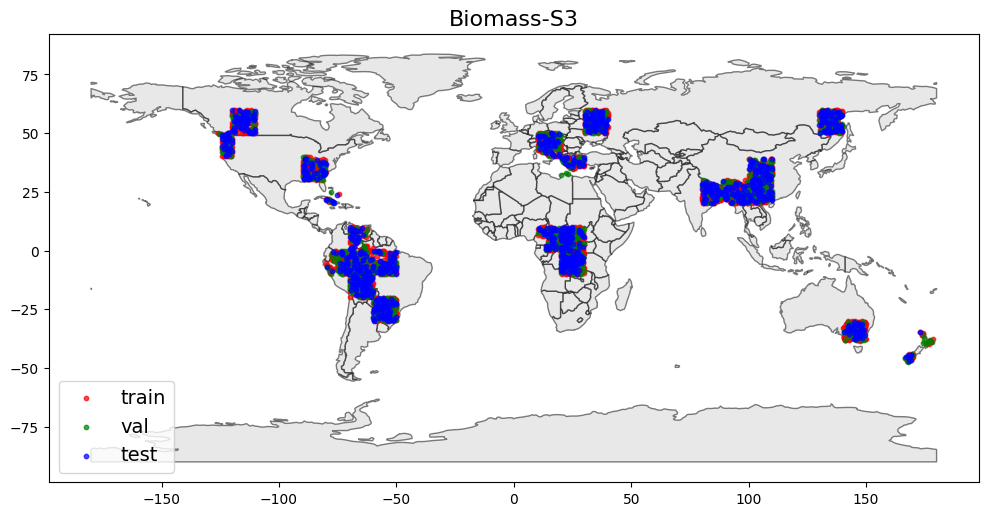}
    \includegraphics[width=0.455\linewidth]{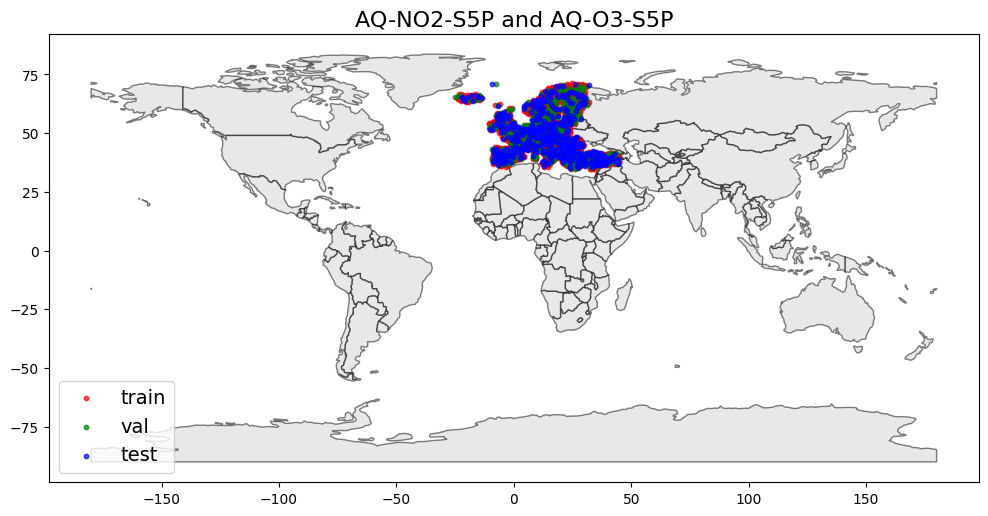}
    
    \caption{Geographical distribution of datasets in Copernicus-Bench.}
    \label{fig:dist}
\end{figure}

\paragraph{Geographical distribution}
\cref{fig:dist} illustrates the geographical distribution of datasets in Copernicus-Bench. Note that DFC2020-S1, DFC2020-S2, and LCZ-S2 do not have geolocation metadata.

\paragraph{Metadata information}

\cref{tab:meta} lists the metadata information of the datasets in Copernicus-Bench.

\begin{table}[ht]
\centering
\caption{Copernicus-Bench metadata availability.
}
\label{tab:meta}
\begin{adjustbox}{max width=\textwidth}
\begin{tabular}{llcccccc}
\toprule
\textbf{Level} & \textbf{Name} & \textbf{Task} & \textbf{Sensor} & \textbf{Bands} & \textbf{Location} & \textbf{Time} & \textbf{Area} \\ \midrule
% \multicolumn{8}{c}{\textbf{Level-1 (Preprocessing)}} \\ \midrule
\multirow{2}{*}{L1} & Cloud-S2 & seg & S2 TOA & All 13 bands & \ding{51} & \ding{51} & \ding{51} \\
 & Cloud-S3 & seg & S3 OLCI & All 21 bands & \ding{51} & \ding{51} & \ding{51} \\ \midrule
% \multicolumn{8}{c}{\textbf{Level-2 (Base Applications)}} \\ \midrule
\multirow{8}{*}{L2} & EuroSAT-S1 & cls & S1 GRD & VV, VH & \ding{51} & \ding{55} & \ding{51} \\
 & EuroSAT-S2 & cls & S2 TOA & All 13 bands & \ding{51} & \ding{55} & \ding{51} \\
 & BigEarthNet-S1 & cls & S1 GRD & VV, VH & \ding{51} & \ding{51} & \ding{51} \\
 & BigEarthNet-S2 & cls & S2 SR & 12 bands (no B10) & \ding{51} & \ding{51} & \ding{51} \\
 & LC100Cls-S3 & cls & S3 OLCI & All 21 bands & \ding{51} & \ding{51} & \ding{51} \\
 & DFC2020-S1 & seg & S1 GRD & VV, VH & \ding{55} & \ding{55} & \ding{51} \\
 & DFC2020-S2 & seg & S2 TOA & All 13 bands & \ding{55} & \ding{55} & \ding{51} \\
 & LC100Seg-S3 & seg & S3 OLCI & All 21 bands & \ding{51} & \ding{51} & \ding{51} \\ \midrule
% \multicolumn{8}{c}{\textbf{Level-3 (Specialized Applications)}} \\ \midrule
\multirow{5}{*}{L3} & Flood-S1 & cd & S1 GRD & VV, VH & \ding{51} & \ding{51} & \ding{51} \\
 & LCZ-S2 & cls & S2 TOA & 10 bands (no B1, B9, B10) & \ding{55} & \ding{55} & \ding{51} \\
 & Biomass-S3 & reg & S3 OLCI & All 21 bands & \ding{51} & \ding{51} & \ding{51} \\
 & AQ-NO2-S5P & reg & S5P NO2 & tropospheric NO\textsubscript{2} column number density & \ding{51} & \ding{51} & \ding{51} \\
 & AQ-O3-S5P & reg & S5P O3 & O\textsubscript{3} column number density & \ding{51} & \ding{51} & \ding{51} \\ \bottomrule
\end{tabular}
\end{adjustbox}
\end{table}

\subsection{Benchmark implementation}

We run all benchmark experiments on a single GPU, repeating three runs with different random seeds. We first benchmark two supervised baselines with ViT-S/16 and ViT-B/16, and then conduct frozen-encoder transfer learning for a set of pretrained models. For classification tasks, a linear layer is appended on top of the encoder; for segmentation and regression tasks, a UPerNet decoder with an auxiliary FCN decoder is appended on top of the encoder; for the flood segmentation task, we follow the segmentation design except that both pre- and post-event images are sent through the encoder and the difference features are sent to the decoder.

We use simplified data augmentations for training sets: horizontal and vertical flipping for classification tasks, and 90$\degree$-rotation, horizontal and vertical flipping for segmentation and regression tasks. No augmentation is used for validation and testing sets. Data normalization is performed on the input according to the pretrained model's preference. For most cases, normalization is performed by subtracting the channel-wise mean and dividing by standard deviation based on the pretrained-model-preferred statistics. If there is no preference, we recommend using the statistics calculated from the training set of each dataset as a standard. For regression tasks, we do mean/std (of the training set) normalization also on the targets to stabilize training. The predicted output is later converted back to the original scale to compute evaluation metrics.

We run each experiment for 50 epochs, and report the test set metrics based on the best validation scores. For classification tasks, we use a batch size of 64, the SGD optimizer, and cross entropy loss or multilabel soft margin loss for single-label or multi-label cases. For segmentation tasks, we use a batch size of 16, the AdamW optimizer, and cross entropy loss. For regression tasks, we use a batch size of 16, the AdamW optimizer, and L1 loss. Specially for air quality regression tasks (NO\textsubscript{2} and O\textsubscript{3}), the targets may contain NaN pixels, thus we customize a masked L1 loss where the NaN pixels do not contribute to the loss calculation. For each model and dataset, we look for the best learning rate with a simple grid search from the pool [1e-4,1e-3,1e-2] (for AdamW) and [1e-2,1e-1, 1, 10] (for SGD). In most cases, the best learning rate is consistent across models but slightly varies across datasets.

\clearpage

%\vspace{2em}
\section{Bridging EO and climate with grid embeddings}
\label{app:climate}

\subsection{Climate prediction visualization}

\begin{figure}[ht]
    \centering
    \includegraphics[width=0.95\linewidth]{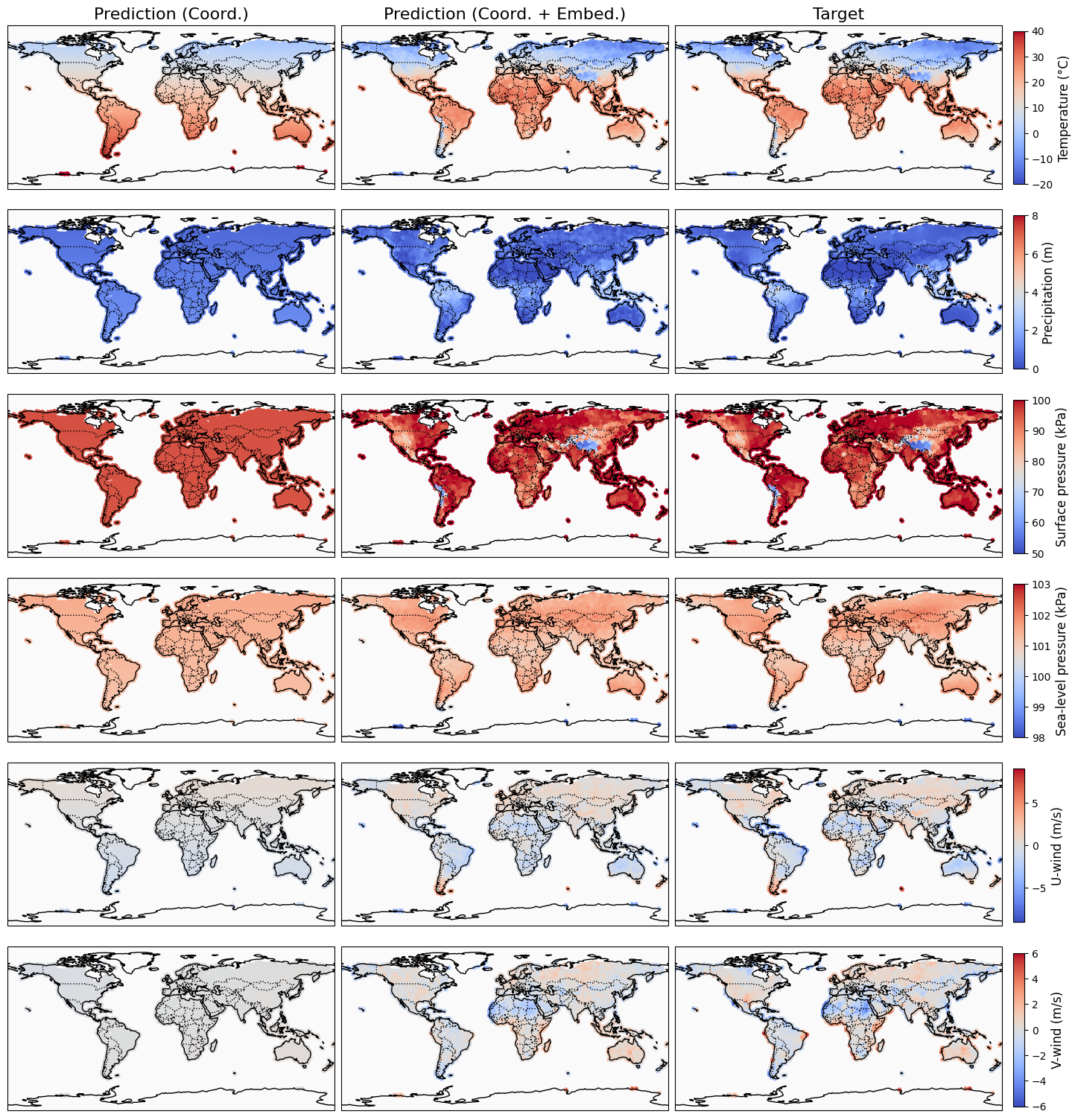}
    \caption{Visualization of climate prediction (10-year mean) results comparing different input sources.}
    \label{fig:climate-pred}
\end{figure}

\begin{figure}[ht]
    \centering
    \includegraphics[width=0.95\linewidth]{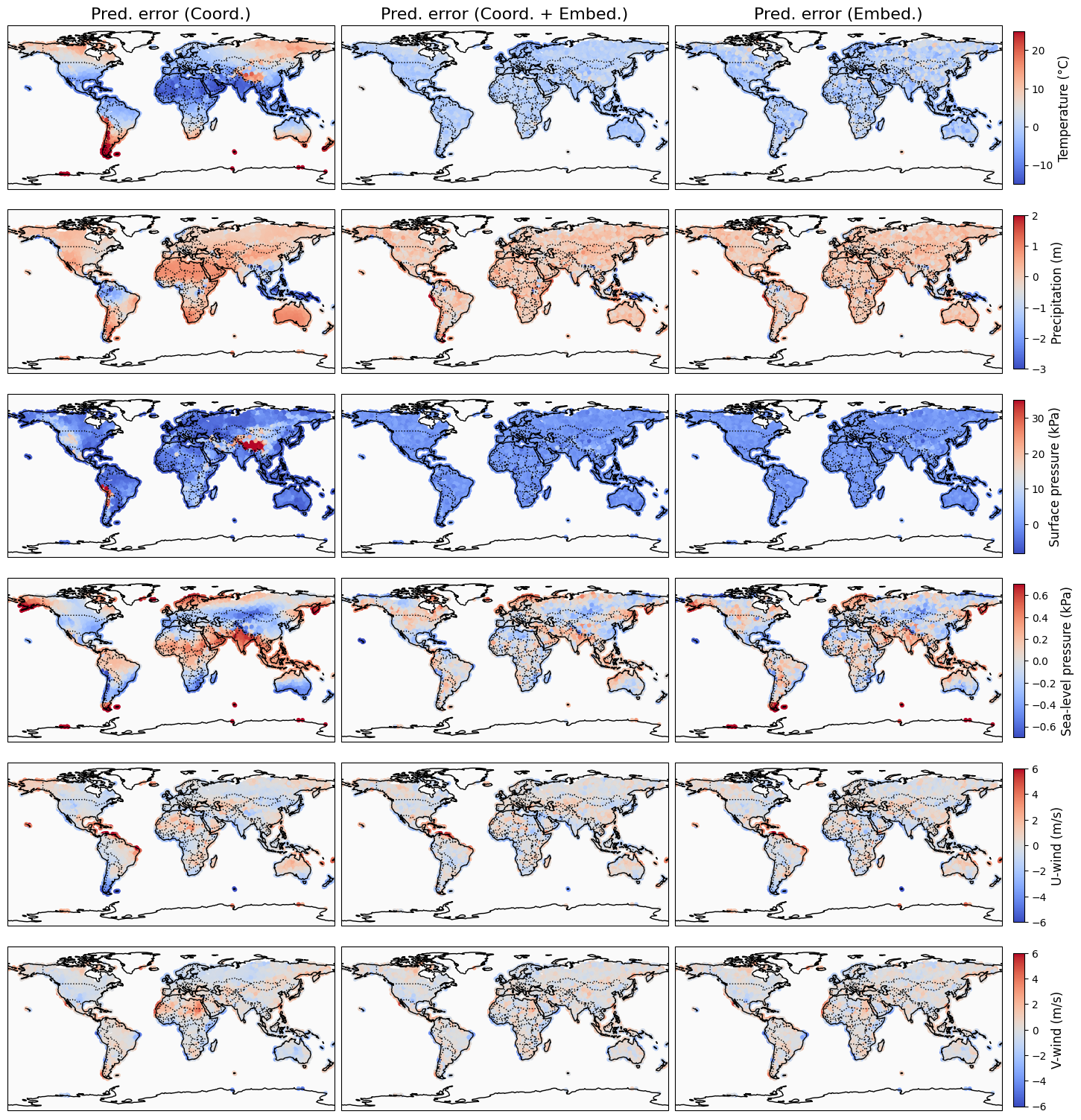}
    \caption{Visualization of climate prediction (10-year mean) errors comparing different input sources.}
    \label{fig:climate-err}
\end{figure}

\begin{figure}[ht]
    \centering
    \includegraphics[width=0.95\linewidth]{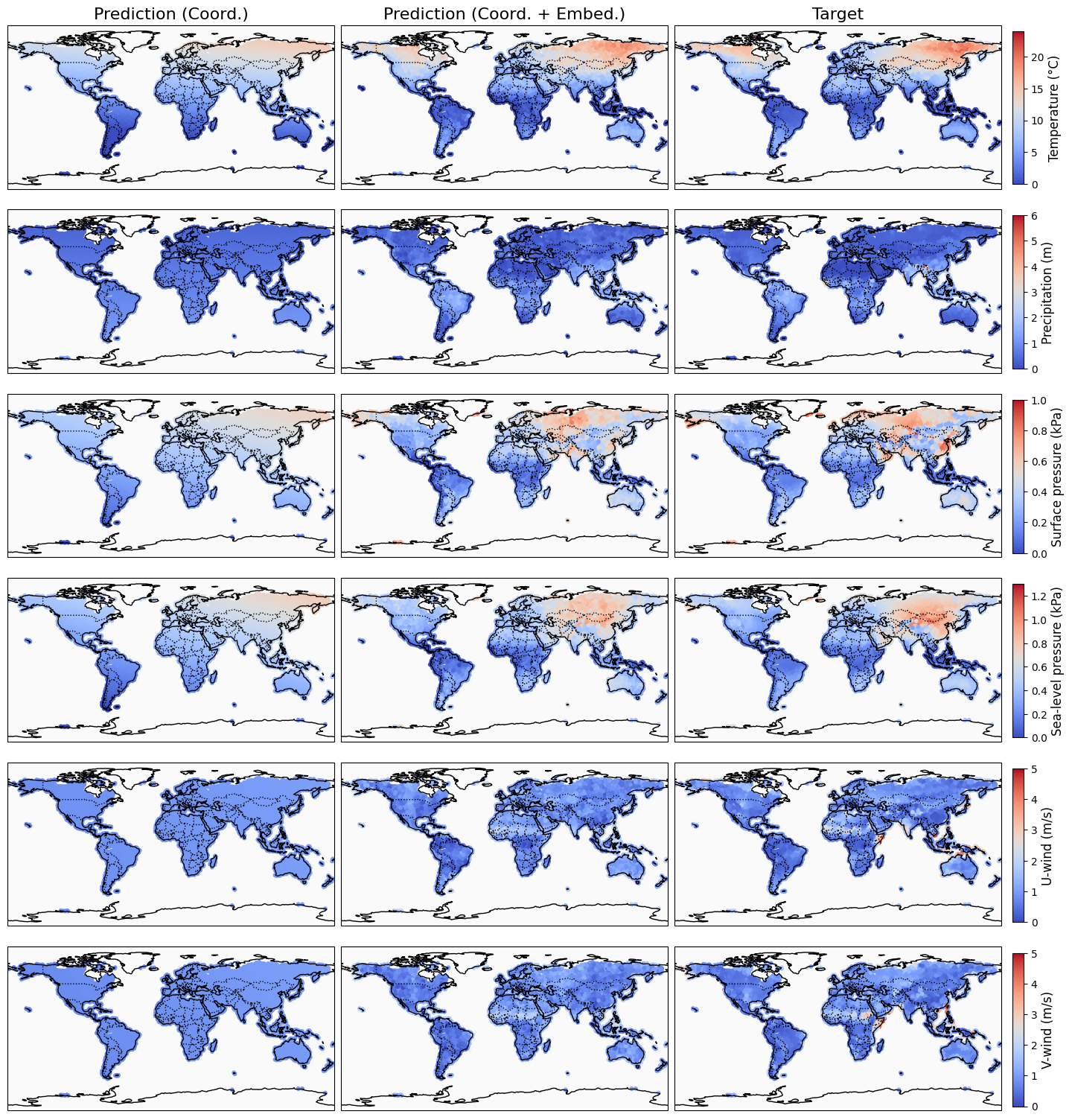}
    \caption{Visualization of climate prediction (10-year std) results comparing different input sources.}
    \label{fig:climate-std-pred}
\end{figure}

\begin{figure}[ht]
    \centering
    \includegraphics[width=0.95\linewidth]{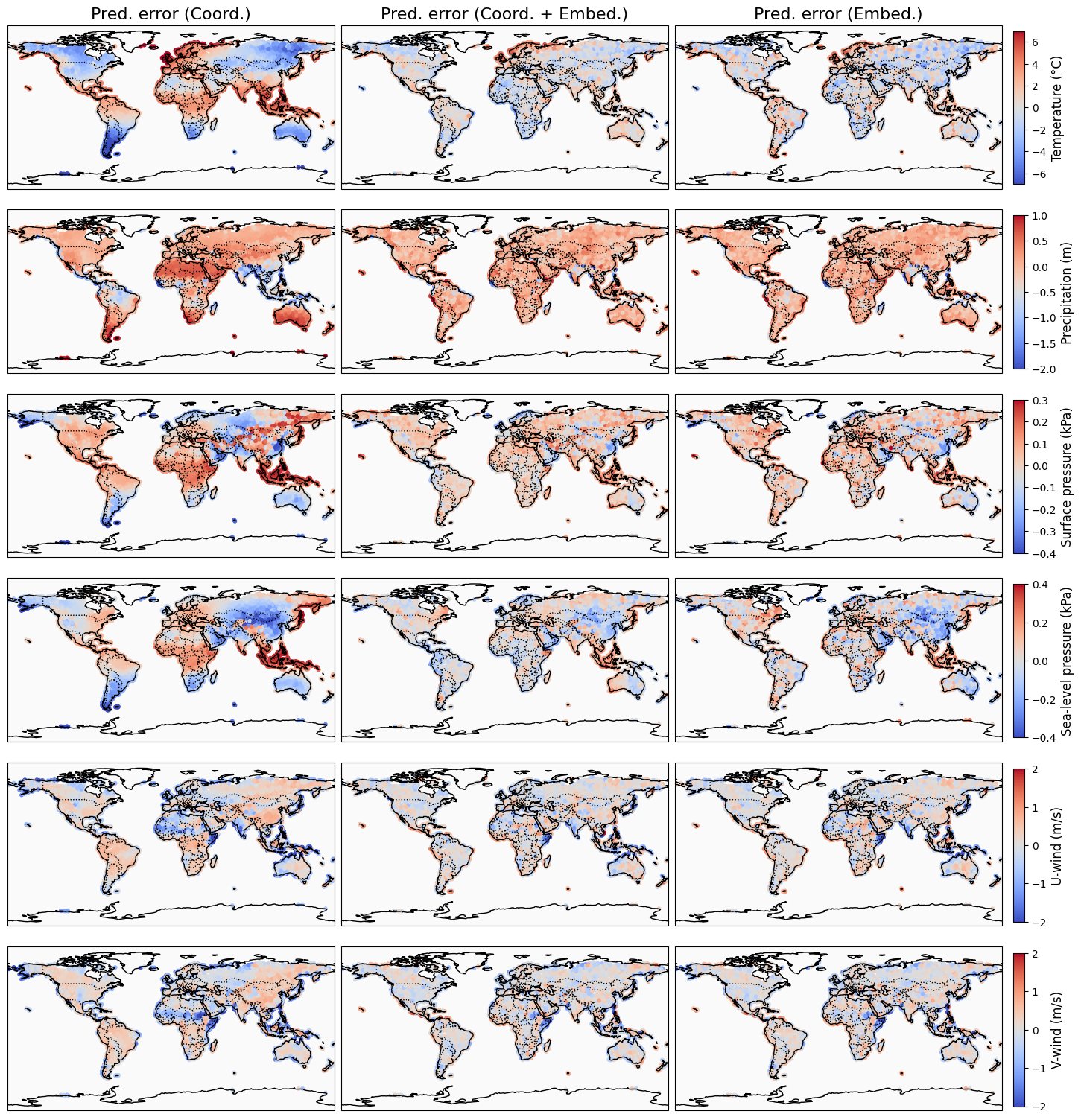}
    \caption{Visualization of climate prediction (10-year std) errors comparing different input sources.}
    \label{fig:climate-std-err}
\end{figure}

\cref{fig:climate-pred,fig:climate-std-pred} visualize the prediction results on 10-year mean/std of the six climate parameters, comparing using the geocoordinates or a combination of coordinates and Copernicus-FM-generated grid embeddings as input data. \cref{fig:climate-err,fig:climate-std-err} further plot the prediction error (Target-Prediction) of using only coordinates, coordinates and embeddings, and only embeddings. The figures show that using raw coordinates captures the general distribution of the climate parameters but tends to be over-smooth, while introducing EO-generated grid embeddings can capture finer details and extremes.

\clearpage

\subsection{Copernicus embedding dataset}

\begin{figure}[ht]
    \centering
    \includegraphics[width=0.95\linewidth]{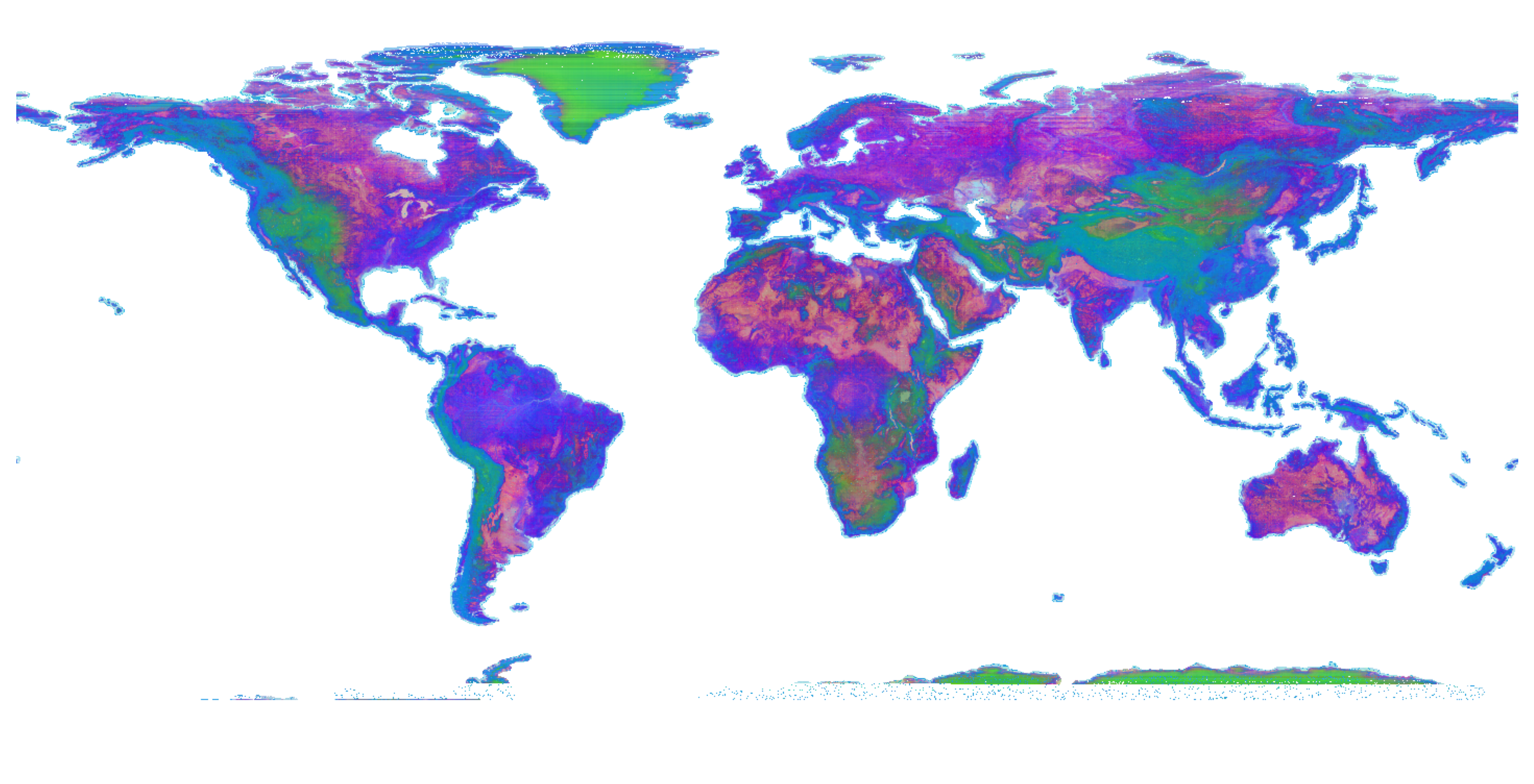}
    \caption{Visualization of the Copernicus-Embed-025deg dataset as a global embedding map (PCA to 3-dim).}
    \label{fig:embed}
\end{figure}

Originally, 10K 0.25$^\circ$$\times$0.25$^\circ$ grids (with all 8 modalities) are sampled from the Copernicus-Pretrain dataset and encoded using the Copernicus-FM model to get image embeddings for each modality. The embeddings are averaged over different modalities to get one embedding vector for each grid, and later used for the climate prediction tasks to investigate the potential of bridging EO and climate. As a follow-up, we extend the embeddings to the whole globe using the full Copernicus-Pretrain dataset, constructing a "global embedding map" at 0.25$^\circ$ with shape 721x1440x768 (filling empty ocean grids with 0). We term this embedding dataset \textbf{Copernicus-Embed-025deg}, which can be seen as a semantic map that integrates various sources of satellite observations at an extremely high compression ratio. This dataset makes it very convenient to link Earth's surface to the atmosphere (e.g., as improved static variables adding to ERA5), unlocking new possibilities in the development of weather/climate foundation models. \cref{fig:embed} visualizes the Copernicus-Embed-025deg dataset with top-3 principal components as RGB channels (empty ocean grids as white background).  

\clearpage
\section{License}
\label{app:license}

All codes, datasets, and model weights will be publicly released under permissive licenses. All codes will be released on GitHub under the Apache-2.0 license, including the curation codes of Copernicus-Pretrain and Copernicus-Bench, the pretraining codes of Copernicus-FM, and the benchmarking codes for Copernicus-Bench. The Copernicus-Pretrain dataset, the newly-curated datasets in Copernicus-Bench, and the pretrained weights of Copernicus-FM will be released under the CC-BY-4.0 license, a copy of which will be hosted on public platforms like Hugging Face. We will also contribute our dataset, model, and benchmark to popular open-source libraries such as TorchGeo~\cite{Stewart_TorchGeo_Deep_Learning_2022}.

\end{document}